%% file: main_arxiv.tex
\documentclass[10pt,twocolumn,letterpaper]{article}

\usepackage{cvpr}
\usepackage{times}
\usepackage{epsfig}
\usepackage{graphicx}
\usepackage{amsmath}
\usepackage{amssymb}

\usepackage{dsfont}
\usepackage{multirow}
\usepackage{booktabs}
\usepackage{algorithm, algorithmic}
\usepackage{rotating}

\newcommand{\cd}{\mathcal{D}}

\newcommand{\cy}{\mathcal{Y}}

\newcommand{\cg}{\mathcal{G}}
\newcommand{\cv}{\mathcal{V}}
\newcommand{\ce}{\mathcal{E}}

\newcommand{\x}{\mathbf{x}}
\newcommand{\y}{\mathbf{y}}
\newcommand{\w}{\mathbf{w}}

\newcommand{\h}{\mathbf{h}}
\newcommand{\m}{\mathbf{m}}

\newcommand{\img}{\mathcal{I}}

\DeclareMathOperator*{\sign}{sign}

\usepackage{pifont}
\newcommand{\cmark}{\ding{51}}%
\newcommand{\xmark}{\ding{55}}%

\usepackage{xr}
\usepackage[pagebackref=true,breaklinks=true,letterpaper=true,colorlinks,bookmarks=false]{hyperref}

\cvprfinalcopy 

\def\cvprPaperID{300} 

\ifcvprfinal\fi
\begin{document}

\title{Learning a Deep ConvNet for Multi-label Classification with Partial Labels}

\author{Thibaut Durand \qquad Nazanin Mehrasa \qquad Greg Mori\\
Borealis AI \qquad Simon Fraser University \\
{\tt\small \{tdurand,nmehrasa\}@sfu.ca} \qquad {\tt\small mori@cs.sfu.ca}
}

\maketitle
\thispagestyle{empty}

\begin{abstract}
Deep ConvNets have shown great performance for single-label image
classification (\eg ImageNet), but it is necessary to move beyond the single-label classification task because pictures of everyday life are inherently multi-label.
Multi-label classification is a more difficult task than single-label
classification because both the input images and output label spaces are more complex.
Furthermore, collecting clean multi-label annotations is more difficult to scale-up than single-label annotations.
To reduce the annotation cost, we propose to train a model with partial labels \ie only some labels are known per image.
We first empirically compare different labeling strategies to show the potential for using partial labels on multi-label datasets.
Then to learn with partial labels, we introduce a new classification
loss that exploits the proportion of known labels per example.
Our approach allows the use of the same training settings as when learning with all the annotations.
We further explore several curriculum learning based strategies to predict missing labels.
Experiments are performed on three large-scale multi-label datasets: MS COCO, NUS-WIDE and Open Images.
\end{abstract}

\input{intro}
\input{sota}
\input{learning}
\input{gnn}
\input{labeling}
\input{experiments}

\input{conclusion}

{\small
\bibliographystyle{ieee}
\bibliography{refs}
}

\clearpage
\appendix
\input{appendix}

\end{document}

%% file: intro.tex
\section{Introduction}

\begin{figure}[t]
\centering
\begin{tabular}{clccc}
\multirow{6}{*}{\includegraphics[height=2.7cm]{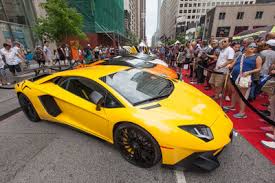}}
&  & [a] & [b] & [c] \\
& \textit{car} & \cmark & \cmark & \cmark \\
& \textit{person} & \cmark &  & \textcolor{red}{\xmark} \\
& \textit{boat} & \xmark &  & \xmark \\
& \textit{bear} & \xmark & \xmark & \xmark \\
& \textit{apple} & \xmark &  & \xmark\\
\\
\end{tabular}
\caption{Example of image with all annotations [a], partial labels [b] and noisy/webly labels [c]. In the partially labeled setting some annotations are missing (person, boat and apple) whereas in the webly labeled setting one annotation is wrong (person).}
\label{fig:intro_web_search}
\end{figure}

Recently, Stock and Cisse \cite{Stock2018} presented empirical evidence that the performance of state-of-the-art classifiers on ImageNet \cite{Russakovsky2015} is largely underestimated -- much of the reamining error is due to the fact that ImageNet's single-label annotation ignores the intrinsic multi-label nature of the images.
Unlike ImageNet, multi-label datasets (\eg MS COCO \cite{Lin2014}, Open Images \cite{Kuznetsova2018}) contain more complex images that represent scenes with several objects (\autoref{fig:intro_web_search}).
However, collecting multi-label annotations is more difficult to scale-up than single-label annotations \cite{Deng2014b}.  As an alternative strategy, one can make use of partial labels; collecting partial labels is easy and scalable with crowdsourcing platforms.
In this work, we study the problem of learning a multi-label classifier with partial labels per image.

The two main (and complementary) strategies to improve image classification performance are: (i) designing / learning better model architectures \cite{Oquab2015, He2016, Sun2016, Zhou2016, Durand2016, Xie2017, Szegedy2017, Durand2017, Pham2018, Zoph2018, Liu2018, Durand2018} and (ii) learning with more labeled data \cite{Sun2017, Mahajan2018}.
However, collecting a multi-label dataset is more difficult and less scalable than collecting a single label dataset \cite{Deng2014b}, because collecting a consistent and exhaustive list of labels for every image requires significant effort.
To overcome this challenge, \cite{Sun2017, Li2017c, Mahajan2018} automatically generated the labels using web supervision.
But the drawback of these approaches is that the annotations are noisy and not exhaustive, and \cite{Zhang2017} showed that learning with corrupted labels can lead to very poor generalization performance.
To be more robust to label noise, some methods have been proposed to learn with noisy labels \cite{Vahdat2017}.

An orthogonal strategy is to use partial annotations.
This direction is actively being pursued by the research community: the largest publicly available multi-label dataset is annotated with partial clean labels \cite{Kuznetsova2018}.
For each image, the labels for some categories are known but the remaining labels are unknown (\autoref{fig:intro_web_search}).
For instance, we know there is a \textit{car} and there is not a \textit{bear} in the image, but we do not know if there is a \textit{person}, a \textit{boat} or an \textit{apple}.  Relaxing the learning requirement for exhaustive labels opens better opportunities for creating large-scale datasets.
Crowdsourcing platforms like Amazon Mechanical Turk\footnote{https://www.mturk.com/} and Google Image Labeler\footnote{https://crowdsource.google.com/imagelabeler/category} or web services like reCAPTCHA\footnote{https://www.google.com/recaptcha/} can scalably collect partial labels for a large number of images.

To our knowledge, this is the first work to examine the challenging task of learning a multi-label image classifier with partial labels on large-scale datasets.
Learning with partial labels on large-scale datasets presents novel challenges because existing methods \cite{Tsoumakas2007, Xu2013, Wu2015, Yang2016} are not scalable and cannot be used to fine-tune a ConvNet.
We address these key technical challenges by introducing a new loss function and a method to fix missing labels.

Our first contribution is to empirically compare several labeling strategies for multi-label datasets to highlight the potential for learning with partial labels.
Given a fixed label budget, our experiments show that partially annotating all images is better than fully annotating a small subset.

As a second contribution, we propose a scalable method to learn a ConvNet with partial labels.
We introduce a loss function that generalizes the standard binary cross-entropy loss by exploiting label proportion information.
This loss automatically adapts to the proportion of known labels per image and allows to use the same training settings as when learning with all the labels.

Our last contribution is a method to predict missing labels.
We show that the learned model is accurate and can be used to predict missing labels.
Because ConvNets are sensitive to noise \cite{Zhang2017}, we propose a curriculum learning based model \cite{Bengio2009} that progressively predicts some missing labels and adds them to the training set.
To improve label predictions, we develop an approach based on Graph Neural Networks (GNNs) to explicitly model the correlation between categories.
In multi-label settings, not all labels are independent, hence reasoning about label correlation between observed and unobserved partial labels is important.

%% file: sota.tex
\section{Related Work}

\paragraph{Learning with partial / missing labels.}
Multi-label tasks often involve incomplete training data, hence several methods have been proposed to solve the problem of multi-label learning with missing labels (MLML).
The first and simple approach is to treat the missing labels as negative labels \cite{Sun2010, Bucak2011, Chen2013b, Wang2014, Sun2017, Mahajan2018}.
The MLML problem then becomes a fully labeled learning problem.
This solution is used in most webly supervised approaches \cite{Sun2017, Mahajan2018}.
The standard assumption is that only the category of the query is present (\eg \textit{car} in \autoref{fig:intro_web_search}) and all the other categories are absent.
However, performance drops because a lot of ground-truth positive labels are initialized as negative labels \cite{Joulin2016}.
A second solution is Binary Relevance (BR) \cite{Tsoumakas2007}, which treats each label as an independent binary classification.
But this approach is not scalable when the number of categories grows and it ignores correlations between labels and between instances, which can be helpful for recognition.
Unlike BR, our proposed approach allows to learn a single model using partial labels.

To overcome the second problem, several works proposed to exploit label correlations from the training data to propagate label information from the provided labels to missing labels.
\cite{Cabral2011, Xu2013} used a matrix completion algorithm to fill in missing labels.
These methods exploit label-label correlations and instance-instance correlations with low-rank regularization on the label matrix to complete the instance-label matrix.
Similarly, \cite{Yu2014} introduced a low rank empirical risk minimization, \cite{Wu2015} used a mixed graph to encode a network of label dependencies and
\cite{Chen2013b,Deng2014b} learned correlation between the categories to predict some missing labels.
Unlike most of the existing models that assume that the correlations are linear and unstructured, \cite{Yang2016} proposed to learn structured semantic correlations.
Another strategy is to treat missing labels as latent variables in probabilistic models.
Missing labels are predicted by posterior inference.
\cite{Kapoor2012, Vasisht2014} used models based on Bayesian networks \cite{Jensen2007} whereas \cite{Chu2018} proposed a deep sequential generative model based on a Variational Auto-Encoder framework \cite{Kingma2014} that also allows to deal with unlabeled data.

However, most of these works cannot be used to learn a deep ConvNet.
They require solving an optimization problem with the training set in memory, so it is not possible to use a mini-batch strategy to fine-tune the model.
This is limiting because it is well-known that fine-tuning is important to transfer a pre-trained architecture \cite{Kornblith2018}.
Some methods are also not scalable because they require to solve convex quadratic optimization problems \cite{Wu2015, Yang2016} that are intractable for large-scale datasets.
Unlike these methods, we propose a model that is scalable and end-to-end learnable.
To train our model, we introduce a new loss function that adapts itself to the proportion of known labels per example.
Similar to some MLML methods, we also explore several strategies to fill-in missing labels by using the learned classifier.

Learning with partial labels is different from semi-supervised learning \cite{Chapelle2010} because in the semi-supervised learning setting, only a subset of the examples is labeled with all the labels and the other examples are unlabeled whereas in the partial labels setting, all the images are labeled but only with a subset of labels.
Note that \cite{Cour2011} also introduced a partially labeled learning problem (also called ambiguously labeled learning) but this problem is different: in \cite{Cour2011}, each example is annotated with multiple labels but only one is correct.

\paragraph{Curriculum Learning / Never-Ending Learning.}
To predict missing labels, we propose an iterative strategy based on Curriculum Learning \cite{Bengio2009}.
The idea of Curriculum Learning is inspired by the way humans learn: start to learn with easy samples/subtasks, and then gradually increase the difficulty level of the samples/subtasks.
But, the main problem in using curriculum learning is to measure the difficulty of an example.
To solve this problem, \cite{Kumar2010} used the definition that easy samples are ones whose correct output can be predicted easily.
They introduced an iterative self-paced learning (SPL) algorithm where each iteration simultaneously selects easy samples and updates the model parameters.
\cite{Jiang2015} generalizes the SPL to different learning schemes by introducing different self-paced functions.
Instead of using human-designed heuristics, \cite{Jiang2018} proposed MentorNet, a method to learn the curriculum from noisy data.
Similar to our work, \cite{Guo2018} recently introduced the CurriculumNet that is a model to learn from large-scale noisy web images with a curriculum learning approach.
However this strategy is designed for multi-class image classification and cannot be used for multi-label image classification because it uses a clustering-based model to measure the difficulty of the examples.

Our approach is also related to the Never-Ending Learning (NEL) paradigm \cite{Mitchell2015}.
The key idea of NEL is to use previously learned knowledge to improve the learning of the model.
\cite{Li2007} proposed a framework that alternatively learns object class models and collects object class datasets.
\cite{Carlson2010, Mitchell2015} introduced the Never-Ending Language Learning to extract knowledge from hundreds of millions of web pages.
Similarly, \cite{Chen2013, Chen2014} proposed the Never-Ending Image Learner to discover structured visual knowledge.
Unlike these approaches that use a previously learned model to extract knowledge from web data, we use the learned model to predict missing labels.

%% file: learning.tex
\section{Learning with Partial Labels}
\label{sec:learning}

Our goal in this paper is to train ConvNets given partial labels.
We first introduce a loss function to learn with partial labels that generalizes the binary cross-entropy.
We then extend the model with a Graph Neural Network to reason about label correlations between observed and unobserved partial labels.
Finally, we use these contributions to learn an accurate model that it is used to predict missing labels with a curriculum-based approach.

\paragraph{Notation.}
We denote by $C$ the number of categories and $N$ the number of training examples.
We denote the training data by $\cd = \{(\img^{(1)}, \y^{(1)}), \ldots ,(\img^{(N)}, \y^{(N)})\}$, where $\img^{(i)}$ is the $i^{th}$ image and $\y^{(i)} = [y^{(i)}_1, \ldots, y^{(i)}_C] \in \cy \subseteq \{-1, 0, 1\}^C$ the label vector.
For a given example $i$ and category $c$, $y^{(i)}_c=1$ (resp.\ $-1$ and $0$) means the category is present (resp.\ absent and unknown).
$\y = [\y^{(1)}; \ldots;\y^{(N)}] \in \{-1, 0, 1\}^{N \times C}$ is
the matrix of training set labels.
$f_\w$ denotes a deep ConvNet with parameters $\w$.
$\x^{(i)} = [x^{(i)}_1, \ldots, x^{(i)}_C] = f_\w(\img^{(i)}) \in \mathbb R^C$ is the output (before sigmoid) of the deep ConvNet $f_\w$ on image $\img^{(i)}$.

\subsection{Binary cross-entropy for partial labels}

The most popular loss function to train a model for multi-label classification is binary cross-entropy (BCE).
To be independent of the number of categories, the BCE loss is normalized by the number of classes.
This becomes a drawback for partially labeled data because the back-propagated gradient becomes small.
To overcome this problem, we propose the partial-BCE loss that normalizes the loss by the proportion of known labels:
\begin{align}
\ell(\x, \y) = \frac{g(p_\y)}{C} \sum_{c=1}^C &\left[ \mathds{1}_{[y_c=1]} \log \left(\frac{1}{1 + \exp(-x_c)}\right) \right. \label{eq:partial_bce} \\
& {} \left. + \mathds{1}_{[y_c=-1]} \log \left(\frac{\exp(-x_c)}{1 + \exp(-x_c)} \right) \right] \nonumber
\end{align}
where $p_\y \in [0, 1]$ is the proportion of known labels in $\y$ and $g$ is a normalization function with respect to the label proportion.
Note that the partial-BCE loss ignores the categories for unknown labels ($y_c=0$).
In the standard BCE loss, the normalization function is $g(p_\y)=1$.
Unlike the standard BCE, the partial-BCE gives the same importance to each example independent of the number of known labels, which is useful when the proportion of labels per image is not fixed.
This loss adapts itself to the proportion of known labels.
We now explain how we design the normalization function $g$.

\begin{figure}[t]
\centering
\includegraphics[width=1\columnwidth]{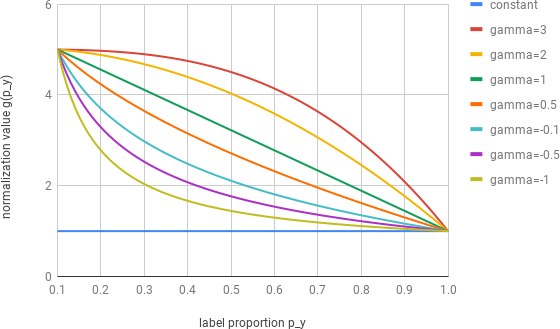}
\caption{Examples of the weight function $g$ (\autoref{eq:norm_function_g}) for different values of hyperparameter $\gamma$ with the constraint $g(0.1) = 5$.
$\gamma$ controls the behavior of the normalization with respect to the label proportion $p_\y$.
}
\label{fig:normalization_function_g}
\end{figure}

\paragraph{Normalization function $g$}.
The function $g$ normalizes the loss function with respect to the label proportion.
We want the partial-BCE loss to have the same behavior as the BCE loss when all the labels are present \ie $g(1) = 1$.
We propose to use the following normalization function:
\begin{align}
g(p_\y) = \alpha p_\y^\gamma + \beta
\label{eq:norm_function_g}
\end{align}
where $\alpha$, $\beta$ and $\gamma$ are the hyperparameters that allow to generalize several standard functions.
For instance with $\alpha=1$, $\beta=0$ and $\gamma=-1$, this function
weights each example inversely proportional to the proportion of
labels.  This is equivalent to normalizing by the number of known classes instead of the number of classes.
Given a $\gamma$ value and the weight for a given proportion (\eg $g(0.1) = 5$), we can find the hyperparameters $\alpha$ and $\beta$ that satisfy these constraints.
The hyperparameter $\gamma$ controls the behavior of the normalization with respect to the label proportion.
In \autoref{fig:normalization_function_g} we show this function for different values of $\gamma$ given the constraint $g(0.1) = 5$.
For $\gamma=1$ the normalization is linearly proportional to the label proportion, whereas for $\gamma=-1$ the normalization value is inversely proportional to the label proportion.
We analyse the importance of each hyperparameter in Sec.\ref{sec:experiments}.
This normalization has a similar goal to batch normalization \cite{Ioffe2015} which normalizes distributions of layer inputs for each mini-batch.

%% file: gnn.tex
\subsection{Multi-label classification with GNN}

To model the interactions between the categories, we use a Graph Neural Network (GNN) \cite{Gori2005, Scarselli2009} on top of a ConvNet.
We first introduce the GNN and then detail how we use GNN for multi-label classification.

\paragraph{GNN.}
For GNNs, the input data is a graph $\cg = \{\cv, \ce\}$ where $\cv$ (resp.\ $\ce$) is the set of nodes (resp.\ edges) of the graph.
For each node $v \in \cv$, we denote the input feature vector $\x_v$ and its hidden representation describing the node's state at time step $t$ by $\h_v^t$.
We use $\Omega_v$ to denote the set of neighboring nodes of $v$.
A node uses information from its neighbors to update its hidden state.
The update is decomposed into two steps: message update and hidden state update.
The message update step combines messages sent to node $v$ into a single message vector $\m_v^t$ according to:
\begin{equation}
\m_v^t = \mathcal{M}(\{\h_u^t | u \in \Omega_v\})
\end{equation}
where $\mathcal{M}$ is the function to update the message.
In the hidden state update step, the hidden states $\h_v^t$ at each node in the graph are updated based on messages $\m_v^t$ according to:
\begin{equation}
\h_v^{t+1} = \mathcal{F}(\h_v^t, \m_v^t)
\end{equation}
where $\mathcal{F}$ is the function to update the hidden state.
$\mathcal{M}$ and $\mathcal{F}$ are feedforward neural networks that are shared among different time steps.
Note that these update functions specify a propagation model of information inside the graph.

\paragraph{GNN for multi-label classification.}
For multi-label classification, each node represents one category ($\cv = \{1, \ldots, C\}$) and the edges represent the connections between the categories.
We use a fully-connected graph to model correlation between all categories.
The node hidden states are initialized with the ConvNet output.
We now detail the GNN functions used in our model.
The algorithm and additional information are given in the
supplementary material.

\paragraph{Message update function $\mathcal{M}$.}
We use the following message update function:
\begin{equation}
\m_v^t = \frac{1}{|\Omega_v| } \sum_{u \in \Omega_v} f_\mathcal{M}(\h_u^t)
\end{equation}
where $f_\mathcal{M}$ is a multi-layer perceptron (MLP).
The message is computed by first feeding hidden states to the MLP $f_\mathcal{M}$ and then taking the average over the neighborhood.

\paragraph{Hidden state update function $\mathcal{F}$.}
We use the following hidden state update function:
\begin{equation}
\h_v^{t+1} = GRU(\h_v^t, \m_v^t)
\end{equation}
which uses a Gated Recurrent Unit (GRU) \cite{Cho2014}.
The hidden state is updated based on the incoming messages and the previous hidden state.

%% file: labeling.tex
\subsection{Prediction of unknown labels}
\label{sec:labeling}

In this section, we propose a method to predict some missing labels with a curriculum learning strategy \cite{Bengio2009}.
We formulate our problem based on the self-paced model \cite{Kumar2010, Jiang2015} and the goal is to optimize the following objective function:
\begin{align}
\underset{\w \in \mathbb R^d, \mathbf{v} \in \{0, 1\}^{N \times C}}{\min} & ~ J(\w, \mathbf{v}) = \beta \|\w\|^2 + G(\mathbf{v}; \theta) \\
& + \frac{1}{N} \sum_{i=1}^N \frac{1}{C} \sum_{c=1}^C v_{ic} \ell_c(f_\w(\img^{(i)}), y_c^{(i)}) \nonumber
\end{align}
where $\ell_c$ is the loss for category $c$ and $v_i \in \{0, 1\}^C$ is a vector to represent the selected labels for the i-th sample. $v_{ic}=1$ (resp.\ $v_{ic}=0$) means that the $c$-th label of the $i$-th example is selected (resp.\ unselected).
The function $G$ defines a curriculum, parameterized by $\theta$, which defines the learning scheme.
Following \cite{Kumar2010}, we use an alternating algorithm where $\w$ and $\mathbf{v}$ are alternatively minimized, one at a time while the other is held fixed. The algorithm is shown in Algorithm \autoref{alg:curriculum}.
Initially, the model is learned with only clean partial labels.
Then, the algorithm uses the learned model to add progressively new ``easy'' weak (\ie noisy) labels in the training set, and then uses the clean and weak labels to continue the training of the model.
We analyze different strategies to add new labels:

\noindent \textbf{[a] Score threshold strategy.}
This strategy uses the classification score (\ie ConvNet) to estimate the difficulty of a pair category-example.
An easy example has a high absolute score whereas a hard example has a score close to 0.
We use the learned model on partial labels to predict the missing labels only if the classification score is larger than a threshold $\theta > 0$.
When $\w$ is fixed, the optimal $\mathbf{v}$ can be derived by:
\begin{align}
v_{ic} = \mathds 1[x_c^{(i)} \geq \theta] + \mathds 1[x_c^{(i)} < -\theta]
\end{align}
The predicted label is $y_c^{(i)} = \sign(x_c^{(i)})$.

\noindent \textbf{[b] Score proportion strategy.}
This strategy is similar to the strategy [a] but instead of labeling the pair category-example higher than a threshold, we label a fixed proportion $\theta$ of pairs per mini-batch.
To find the optimal $\mathbf{v}$, we sort the examples by decreasing order of absolute score and label only the top-$\theta$\% of the missing labels.

\noindent \textbf{[c] Predict only positive labels.}
Because of the imbalanced annotations, we only predict positive labels with strategy [a].
When $\w$ is fixed, the optimal $\mathbf{v}$ can be derived by:
\begin{align}
v_{ic} = \mathds 1[x_c^{(i)} \geq \theta]
\end{align}

\noindent \textbf{[d] Ensemble score threshold strategy.}
This strategy is similar to the strategy [a] but it uses an ensemble of models to estimate the confidence score.
We average the classification score of each model to estimate the final confidence score.
This strategy allows to be more robust than the strategy [a].
When $\w$ is fixed, the optimal $\mathbf{v}$ can be derived by:
\begin{align}
v_{ic} = \mathds 1[E(\img^{(i)})_c \geq \theta] + \mathds 1[E(\img^{(i)})_c < -\theta]
\end{align}
where $E(\img^{(i)}) \in \mathbb R^C$ is the vector score of an ensemble of models.
The predicted label is $y_c^{(i)} = \sign(E(\img^{(i)})_c)$.

\noindent \textbf{[e] Bayesian uncertainty strategy.}
Instead of using the classification score as in [a] or [d], we estimate the bayesian uncertainty \cite{Kendall2017} of each pair category-example. An easy pair category-example has a small uncertainty.
When $\w$ is fixed, the optimal $\mathbf{v}$ can be derived by:
\begin{align}
v_{ic} = \mathds 1[U(\img^{(i)})_c \leq \theta]
\end{align}
where $U(\img^{(i)})$ is the bayesian uncertainty of category $c$ of the $i$-th example.
This strategy is similar to strategy [d] except that it uses the variance of the classification scores instead of the average to estimate the difficulty.

\begin{algorithm}[t]
\caption{Curriculum labeling}
\label{alg:curriculum}
\begin{algorithmic}[1]
\REQUIRE Training data $\cd$
\STATE Initialize $\mathbf{v}$ with known labels
\STATE Initialize $\w$: learn the ConvNet with the partial labels
\REPEAT
\STATE Update $\mathbf{v}$ (fixed $\w$): find easy missing labels
\STATE Update $\y$: predict the label of easy missing labels
\STATE Update $\w$ (fixed $\mathbf{v}$): improve classification model with the clean and easy weak annotations
\UNTIL{stopping criteria}
\end{algorithmic}
\end{algorithm}

%% file: experiments.tex
\section{Experiments}
\label{sec:experiments}

\paragraph{Datasets.}
We perform experiments on several standard multi-label datasets: Pascal VOC 2007 \cite{Everingham2015}, MS COCO \cite{Lin2014} and NUS-WIDE \cite{Chua2009}.
For each dataset, we use the standard train/test sets introduced respectively in \cite{Everingham2015}, \cite{Oquab2014}, and \cite{Gong2014} (see \autoref{sec:app_exp_details} of supplementary for more details).
From these datasets that are fully labeled, we create partially labeled datasets by randomly dropping some labels per image.
The proportion of known labels is between 10\% (90\% of labels
missing) and 100\% (all labels present).
We also perform experiments on the large-scale Open Images dataset \cite{Kuznetsova2018} that is partially annotated: $0.9\%$ of the labels are available during training.

\paragraph{Metrics.}
To evaluate the performances, we use several metrics:
mean Average Precision (MAP) \cite{BaezaYates1999}, 0-1 exact match, Macro-F1 \cite{Yang1999}, Micro-F1 \cite{Tang2009}, per-class precision, per-class recall, overall precision, overall recall.
These metrics are standard multi-label classification metrics and are presented in \autoref{sec:app_metrics} of supplementary.
We mainly show the results for the MAP metric but results for other metrics are shown in supplementary.

\paragraph{Implementation details.}
We employ ResNet-WELDON \cite{Durand2018} as our classification network.
We use a ResNet-101 \cite{He2016} pretrained on ImageNet as the backbone architecture, but we show results for other architectures in supplementary.
The models are implemented with PyTorch \cite{Paszke2017}.
The hyperparameters of the partial-BCE loss function are $\alpha=-4.45$, $\beta=5.45$ (\ie $g(0.1)=5$) and $\gamma=1$.
To predict missing labels, we use the bayesian uncertainty strategy with $\theta=0.3$.

\subsection{What is the best strategy to annotate a dataset?}

\begin{figure*}[t]
\centering
\begin{tabular}{ccc}
\includegraphics[width=0.65\columnwidth]{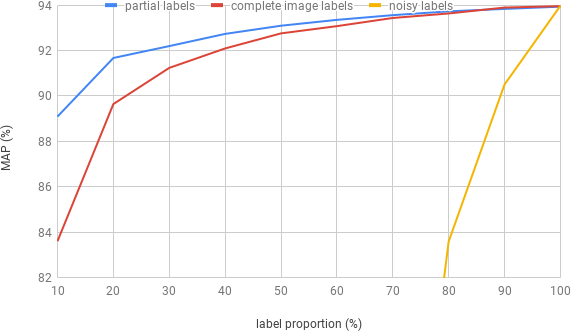} &
\includegraphics[width=0.65\columnwidth]{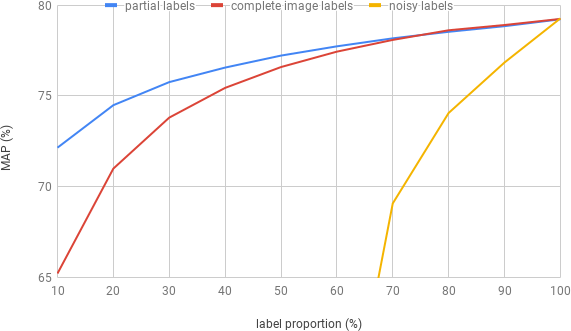} &
\includegraphics[width=0.65\columnwidth]{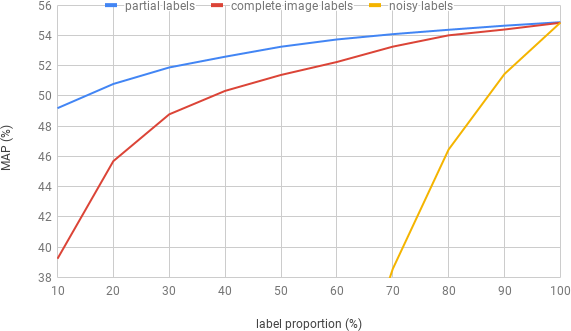}
\\
\includegraphics[width=0.65\columnwidth]{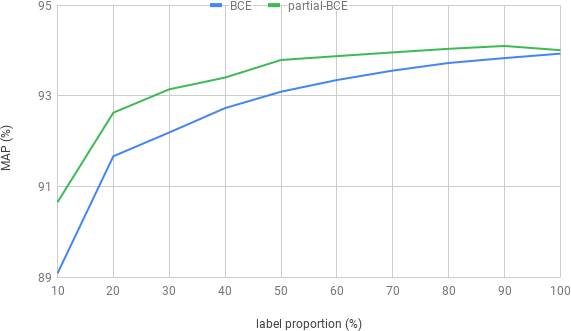} &
\includegraphics[width=0.65\columnwidth]{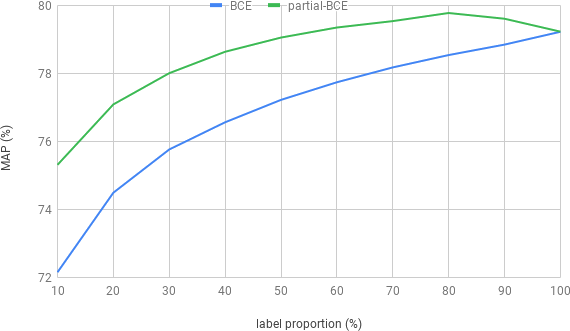} &
\includegraphics[width=0.65\columnwidth]{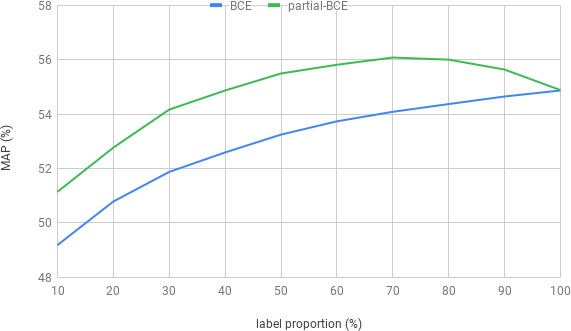}
\\
Pascal VOC 2007 & MS COCO & NUS-WIDE
\end{tabular}
\caption{The first row shows MAP results for the different labeling strategies. On the second row, we shows the comparison of the BCE and the partial-BCE. The x-axis shows the proportion of clean labels.}
\label{fig:partial_results}
\end{figure*}

In the first set of experiments, we study three strategies to annotate a multi-label dataset.
The goal is to answer the question: what is the best strategy to annotate a dataset with a fixed budget of clean labels?
We explore the three following scenarios:
\begin{itemize}
\item \textbf{Partial labels}.
This is the strategy used in this paper.
In this setting, all the images are used but only a subset of the labels per image are known.
The known categories are different for each image.

\item \textbf{Complete image labels or dense labels}.
In this scenario, only a subset of the images are labeled, but the labeled images have the annotations for all the categories.
This is the standard setting for semi-supervised learning \cite{Chapelle2010} except that we do not use a semi-supervised model.
\item \textbf{Noisy labels}.
All the categories of all images are labeled but some labels are wrong.
This scenario is similar to the webly-supervised learning scenario \cite{Mahajan2018} where some labels are wrong.
\end{itemize}
To have fair comparison between the approaches, we use a BCE loss function for these experiments.
The results are shown in \autoref{fig:partial_results} for different proportion of clean labels.
For each experiment, we use the same number of clean labels.
$100\%$ means that all the labels are known during training (standard classification setting) and $10\%$ means that only $10\%$ of the labels are known during training.
The 90\% of other labels are unknown labels for the partial labels and the complete image labels scenarios and are wrong labels for the noisy labels scenario.
Similar to \cite{Sun2017}, we observe that the performance increases logarithmically based on proportion of labels.
From this first experiment, we can draw the following conclusions:
(1) Given a fixed number of clean labels, we observe that learning with partial labels is better than learning with a subset of dense annotations.
The improvement increases when the label proportion decreases.
A reason is that the model trained in the partial labels strategy ``sees'' more images during training and therefore has a better generalization performance.
(2) It is better to learn with a small subset of clean labels than a lot of labels with some incorrect labels.
Both partial labels and complete image labels scenarios are better than the noisy label scenario.
For instance on MS COCO, we observe that learning with only 20\% of clean partial labels is better than learning with 80\% of clean labels and 20\% of wrong labels.

\paragraph{Noisy web labels.}
Another strategy to generate a noisy dataset from a multi-label dataset is to use only one positive label for each image.
This is a standard assumption made when collecting data from the web \cite{Li2017c} \ie the only category present in the image is the category of the query.
From the clean MS COCO dataset, we generate a noisy dataset (named noisy+) by keeping only one positive label per image.
If the image has more than one positive label, we randomly select one positive label among the positive labels and switch the other positive labels to negative labels.
The results are reported in \autoref{tab:expes_multi_single} for three scenarios: clean (all the training labels are known and clean), 10\% of partial labels and noisy+ scenario.
We also show the percentage of clean and noisy labels for each experiment.
The noisy+ approach generates a small proportion of noisy labels (2.4\%) that  drops the performance by about 7pt with respect to the clean baseline.
We observe that a model trained with only 10\% of clean labels is slightly better than the model trained with the noisy labels.
This experiment shows that the standard assumption made in most of the webly-supervised datasets is not good for complex scenes / multi-label images because it generates noisy labels that significantly decrease generalization.

\begin{table}[t]
\centering
\begin{tabular}{lccc}
\toprule
model & clean & partial 10\% & noisy+ \\
\midrule
clean / noisy labels & 100 / 0 & 10 / 0 & 97.6 / 2.4 \\
MAP (\%) & 79.22 & 72.15 & 71.60  \\
\bottomrule
\end{tabular}
\caption{Comparison with a webly-supervised strategy (noisy+) on MS COCO.
Clean (resp. noisy) means the percentage of clean (resp. noisy) labels in the training set.}
\label{tab:expes_multi_single}
\end{table}

\begin{table*}[ht]
\centering
\begin{tabular}{lcccc|cccc}
\toprule
Relabeling & MAP & 0-1 & Macro-F1 & Micro-F1 & label prop. & TP & TN & GNN \\
\midrule
2 steps (no curriculum) & -1.49 & 6.42 & 2.32 & 1.99 & 100 & 82.78 & 96.40 & \cmark \\
\midrule
\textbf{[a]} Score threshold $\theta=2$ & 0.34 & 11.15 & 4.33 & 4.26 & 95.29 & 85.00 & 98.50 & \cmark \\
\textbf{[b]} Score proportion $\theta=80\%$ & 0.17 & 8.40 & 3.70 & 3.25 & 96.24 & 84.40 & 98.10 & \cmark \\
\textbf{[c]} Postitive only - score $\theta=5$ & 0.31 & -4.58 & -1.92 & -2.23 & 12.01 & 79.07 & - & \cmark \\
\textbf{[d]} Ensemble score $\theta=2$ & 0.23 & 11.31 & 4.16 & 4.33 & 95.33 & 84.80 & 98.53 & \cmark \\
\midrule
\textbf{[e]} Bayesian uncertainty $\theta=0.3$ & 0.34 & 10.15 & 4.37 & 3.72 & 77.91 & 61.15 & 99.24 & \\
\midrule
\textbf{[e]} Bayesian uncertainty $\theta=0.1$ & 0.36 & 2.71 & 1.91 & 1.22 & 19.45 & 38.15 & 99.97 & \cmark \\
\textbf{[e]} Bayesian uncertainty $\theta=0.2$ & 0.30 & 10.76 & 4.87 & 4.66 & 57.03 & 62.03 & 99.65 & \cmark \\
\textbf{[e]} Bayesian uncertainty $\theta=0.3$ & 0.59 & 12.07 & 5.11 & 4.95 & 79.74 & 68.96 & 99.23 & \cmark \\
\textbf{[e]} Bayesian uncertainty $\theta=0.4$ & 0.43 & 10.99 & 4.88 & 4.46 & 90.51 & 70.77 & 98.57 & \cmark \\
\textbf{[e]} Bayesian uncertainty $\theta=0.5$ & 0.45 & 10.08 & 3.93 & 3.78 & 94.79 & 74.73 & 98.00 & \cmark \\
\bottomrule
\end{tabular}
\caption{Analysis of the labeling strategy of missing labels on Pascal VOC 2007 val set.
For each metric, we report the relative scores with respect to a model that does not label missing labels.
TP (resp. TN) means true positive (resp. true negative) rate.
For the strategy [c], we report the label accuracy instead of the TP rate. }
\label{tab:expes_relabeling_voc2007}
\end{table*}

\subsection{Learning with partial labels}

In this section, we compare the standard BCE and the partial-BCE and analyze the importance of the GNN.

\paragraph{BCE vs partial-BCE.}
The \autoref{fig:partial_results} shows the MAP results for different proportion of known labels on three datasets.
For all the datasets, we observe that using the partial-BCE significantly improves the performance: the lower the label proportion, the better the improvement.
We observe the same behavior for the other metrics (\autoref{sec:app_compa_loss} of supplementary).
In \autoref{tab:open_images}, we show results on the Open Images dataset and we observe that the partial-BCE is 4 pt better than the standard BCE.
These experiments show that our loss learns better than the BCE because it exploits the label proportion information during training.
It allows to learn efficiently while keeping the same training setting as with all annotations.

\begin{table}[t]
\centering
\begin{tabular}{lccccc}
\toprule
 & BCE & partial-BCE & GNN + partial-BCE \\
\midrule
MAP (\%) & 79.01 & 83.05 & 83.36 \\
\bottomrule
\end{tabular}
\caption{MAP results on Open Images.}
\label{tab:open_images}
\end{table}

\vspace{-2mm}

\begin{figure}[t]
\centering
\includegraphics[width=1\columnwidth]{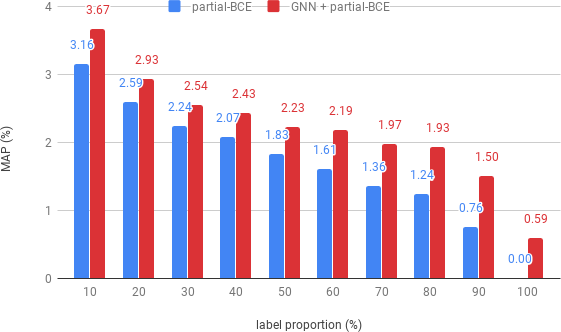}
\caption{MAP (\%) improvement with respect to the proportion of known labels on MS COCO for the partial-BCE and the GNN + partial-BCE. 0 means the result for a model trained with the standard BCE.}
\label{fig:expes_coco_gnn}
\end{figure}

\paragraph{GNN.}
We now analyze the improvements of the GNN to learn relationships between the categories.
We show the results on MS COCO in \autoref{fig:expes_coco_gnn}.
We observe that for each label proportion, using the GNN improves the performance.
Open Images experiments (\autoref{tab:open_images}) show that GNN improves the performance even when the label proportion is small.
This experiment shows that modeling the correlation between categories is important even in case of partial labels.
However, we also note that a ConvNet implicitly learns some correlation between the categories because some learned representations are shared by all categories.

\begin{table*}
\centering
\begin{tabular}{ccccc|cccccc}
\toprule
BCE & fine-tuning & partial-BCE & GNN & relabeling & MAP & 0-1 exact match & Macro-F1 & Micro-F1 \\
\midrule
\cmark & & & & & 66.21 & 17.53 & 62.74 & 67.33 \\
\cmark & \cmark & & & & 72.15 & 22.04 & 65.82 & 70.09 \\
& \cmark & \cmark & & & 75.31 & 24.51 & 67.94 & 71.18 \\
& \cmark & \cmark & \cmark & & 75.82 & 25.14 & 68.40 & 71.37 \\
& \cmark & \cmark &  & \cmark & 75.71 & 30.52 & 70.13 & 73.87 \\
& \cmark & \cmark & \cmark & \cmark & 76.40 & 32.12 & 70.73 & 74.37 \\
\bottomrule
\end{tabular}
\caption{Ablation study on MS COCO with 10\% of known labels.}
\label{tab:ablation_study_coco}
\end{table*}

\subsection{What is the best strategy to predict missing labels?}

In this section, we analyze the labeling strategies introduced in \autoref{sec:labeling} to predict missing labels.
Before training epochs 10 and 15, we use the learned classifier to predict some missing labels.
We report the results for different metrics on Pascal VOC 2007 validation set with 10\% of labels in \autoref{tab:expes_relabeling_voc2007}.
We also report the final proportion of labels, the true postive (TP) and true negative (TN) rates for predicted labels.
Additional results are shown in \autoref{sec:app_label_missing} of supplementary.

First, we show the results of a 2 steps strategy that predicts all missing labels in one time.
Overall, we observe that this strategy is worse than curriculum-based strategies
([a-e]).
In particular, the 2 steps strategy decreases the MAP score.
These results show that predicting all missing labels at once introduced too much label noise, decreasing generalization performance.
Among the curriculum-based strategies, we observe that the threshold strategy [a] is better than the proportion strategy [b].
We also note that using a model ensemble [d] does not significantly improve the performance with respect to a single model [a].
Predicting only positive labels [c] is a poor strategy.
The bayesian uncertainty strategy [e] is the best strategy.
In particular, we observe that the GNN is important for this strategy because it decreases the label uncertainty and allows the model to be robust to the hyperparameter $\theta$.

\input{experiments_loss_analysis}

%% file: experiments_loss_analysis.tex
\subsection{Method analysis}

In this section, we analyze the hyperparameters of the partial-BCE and perform an ablation study on MS COCO.

\begin{figure}[t]
\centering
\includegraphics[width=1\columnwidth]{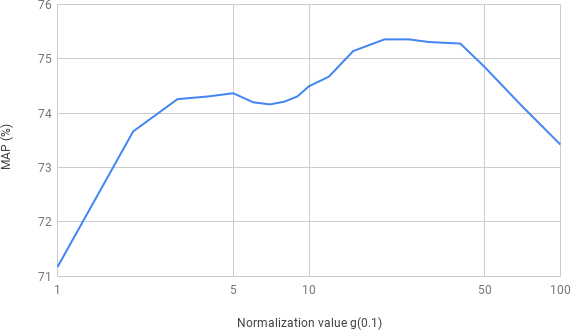}
\caption{Analysis of the normalization value  for a label proportion of 10\% (\ie $g(0.1)$). (x-axis log-scale)}
\label{fig:expes_weight_y}
\end{figure}

\paragraph{Partial-BCE analysis.}
To analyze the partial-BCE, we use only the training set.
The model is trained on about 78k images and evaluated on the remaining 5k images.
We first analyse how to choose the value of the normalization function given a label proportion of 10\% \ie $g(0.1)$ (it is possible to choose another label proportion).
The results are shown in \autoref{fig:expes_weight_y}.
Note that for $g(0.1)=1$, the partial-BCE is equivalent to the BCE and the loss is normalized by the number of categories.
We observe that the normalization value $g(0.1)=1$ gives the worst results.
The best score is obtained for a normalization value around 20 but the performance is similar for $g(0.1) \in [3, 50]$.
Using a large value drops the performance.
This experiment shows that the proposed normalization function is important and robust.
These results are independent of the network architectures (\autoref{sec:app_loss_analysis} of supplementary).

Given the constraints $g(0.1)=5$ and $g(1)=1$, we analyze the impact of the hyperparameter $\gamma$.
This hyperparameter controls the behavior of the normalization with respect to the label proportion.
Using a high value ($\gamma=3$) is better than a low value ($\gamma=-1$) for large label proportions but is slighty worse for small label proportions.
We observe that using a normalization that is proportional to the number of known labels ($\gamma=1$) works better than using a normalization that is inversely proportional to the number of known labels ($\gamma=-1$).

\begin{figure}[t]
\centering
\includegraphics[width=1\columnwidth]{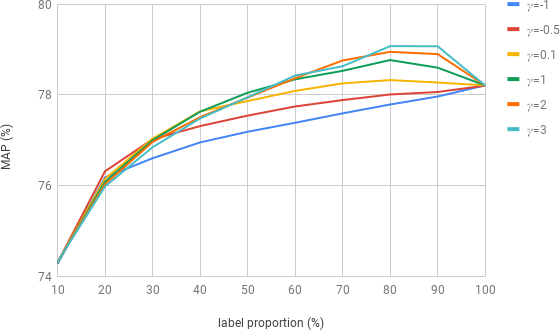}
\caption{Analysis of hyperparameter $\gamma$ on MS COCO.}
\label{fig:expes_weight_gamma}
\end{figure}

\paragraph{Ablation study.}
Finally to analyze the importance of each contribution, we perform an ablation study on MS COCO for a label proportion of 10\% in \autoref{tab:ablation_study_coco}.
We first observe that fine-tuning is important.
It validates the importance of building end-to-end trainable models to learn with missing labels.
The partial-BCE loss function increases the performance against each metric because it exploits the label proportion information during training.
We show that using GNN or relabeling improves performance.
In particular, the relabeling stage significantly increases the 0-1 exact match score (+5pt) and the Micro-F1 score (+2.5pt).
Finally, we observe that our contributions are complementary.

%% file: conclusion.tex
\section{Conclusion}

In this paper, we present a scalable approach to end-to-end learn a multi-label classifier with partial labels.
Our experiments show that our loss function significantly improves performance.
We show that our curriculum learning model using bayesian uncertainty is an accurate strategy to label missing labels.
In the future work, one could combine several datasets whith shared categories to learn with more training data.

%% file: appendix.tex
\section{Supplementary}

\input{appendix_gnn}

\subsection{Experimental details}
\label{sec:app_exp_details}

\paragraph{Datasets.}
We perform experiments on large publicly available multi-label datasets: Pascal VOC 2007 \cite{Everingham2015}, MS COCO \cite{Lin2014} and NUS-WIDE \cite{Chua2009}.
Pascal VOC 2007 dataset contains 5k/5k trainval/test images of 20 objects categories.
MS COCO dataset contains 123k images of 80 objects categories.
We use the 2014 data split with 83k train images and 41k val images.
NUS-WIDE dataset contains 269,648 images downloaded from Flickr that have been manually annotated with 81 visual concepts.
We follow the experimental protocol in \cite{Gong2014} and use 150k randomly sampled images for training and the rest for testing.
The results on NUS-WIDE cannot be directly comparable with the other works because the number of total images is different (209,347 in \cite{Gong2014}, 200,261 in \cite{Li2017b}).
The main reason is that some provided URLs are invalid or some images have been deleted from Flickr.
For our experiments, we collected 216,450 images.

We also performs experiments on the largest publicly available multi-label dataset: Open Images \cite{Kuznetsova2018}.
This dataset is partially annotated with human labels and machine generated labels.
For our experiments, we use only human labels on the 600 boxable classes.
On the training set, only $0.9\%$ of the labels are available.

\paragraph{Implementation details.}
The hyperparameters of the WELDON pooling function are $k^+ = k^- = 0.1$.
The models are implemented with PyTorch \cite{Paszke2017} and are trained with SGD during 20 epochs with a batch size of 16.
The initial learning rate is $0.01$ and it is divide by 10 after 10 epochs.
During training, we only use random horizontal flip as data augmentation.
Each image is resized to $448 \times 448$ with 3 color channels.
On Open Images dataset, unlike \cite{Kuznetsova2018} we do not train from scratch the network.
We use a similar protocol that on the others datasets: we fine-tune a model pre-train on ImageNet but stop the training when the validation performance does not increase. Because the training set has 1.7M images, the model converge in less than 5 epochs.

\subsection{Multi-label metrics}
\label{sec:app_metrics}

In this section, we introduce the metrics used to evaluate the performances on multi-label datasets.
We note $\y^{(i)} = [y^{(i)}_1, \ldots, y^{(i)}_C] \in \cy \subseteq \{-1, 0, 1\}^C$ the ground truth label vector and $\hat{\y}^{(i)} = [\hat{y}^{(i)}_1, \ldots, \hat{y}^{(i)}_C] \in \{-1, 1\}^C$ the predicted label vector of the $i$-th example.

\paragraph{Zero-one exact match accuracy (0-1).}
This metric considers a prediction correct only if all the labels are correctly predicted:
\begin{align}
m_{0/1}(\cd) = \frac{1}{N} \sum_{i=1}^N \mathds 1[\y^{(i)} = \hat{\y}^{(i)}]
\end{align}
where $\mathds 1[.]$ is an indicator function.

\paragraph{Per-class precision/recall (PC-P/R).}
\begin{align}
m_{PC-P}(\cd) &= \frac{1}{C} \sum_{c=1}^C \frac{N_c^{correct}}{N_c^{predict}}
\\
m_{PC-R}(\cd) &= \frac{1}{C} \sum_{c=1}^C \frac{N_c^{correct}}{N_c^{gt}}
\end{align}
where $N_c^{correct}$ is the number of correctly predicted images for the $c$-th label, $N_c^{predict}$ is the number of predicted images, $N_c^{gt}$ is the number of ground-truth images.
Note that the per-class measures treat all classes equal regardless of their sample size, so one can obtain a high performance by focusing on getting rare classes right.

\paragraph{Overall precision/recall (OV-P/R).}
Unlike per-class metrics, the overall metrics treat all samples equal regardless of their classes.
\begin{align}
m_{OV-P}(\cd) &= \frac{\sum_{c=1}^C N_c^{correct}}{\sum_{c=1}^C N_c^{predict}}
\\
m_{OV-R}(\cd) &= \frac{\sum_{c=1}^C N_c^{correct}}{\sum_{c=1}^C N_c^{gt}}
\end{align}

\paragraph{Macro-F1 (M-F1).} The macro-F1 score \cite{Yang1999} is the F1 score \cite{Rijsbergen1979} averaged across all categories.
\begin{align}
m_{MF1}(\cd) &= \frac{1}{C} \sum_{c=1}^C F_1^c
\end{align}
Given a category $c$, the F1 measure, defined as the harmonic mean of precision and recall, is computed as follows:
\begin{align}
F_1^c = \frac{2 P^c R^c}{P^c + R^c}
\end{align}
where the precision ($P^c$) and the recall ($R^c$) are calculated as follows:
\begin{align}
P^c = \frac{\sum_{i=1}^N \mathds 1[y_c^{(i)} = \hat{y}_c^{(i)}]}{\sum_{i=1}^N \hat{y}_c^{(i)}}
\\
R^c = \frac{\sum_{i=1}^N \mathds 1[y_c^{(i)} = \hat{y}_c^{(i)}]}{\sum_{i=1}^N y_c^{(i)}}
\end{align}
and $y_c^{(i)} \in \{0, 1\}$

\paragraph{Micro-F1 (m-F1).}
The micro-F1 score \cite{Tang2009} is computed using the equation of $F_1^c$ and considering
the predictions as a whole
\begin{align}
m_{mF1}(\cd) &= \frac{2 \sum_{c=1}^C \sum_{i=1}^N \mathds 1[y_c^{(i)} = \hat{y}_c^{(i)}]}{\sum_{c=1}^C \sum_{i=1}^N y_c^{(i)} + \sum_{c=1}^C \sum_{i=1}^N \hat{y}_c^{(i)} }
\end{align}
According to the definition, macro-F1 is more sensitive to the performance of rare categories while micro-F1 is affected more by the major categories.

\subsection{Analysis of the initial set of labels}

\begin{figure*}[t]
\centering
\begin{tabular}{cc}
\includegraphics[width=1\columnwidth]{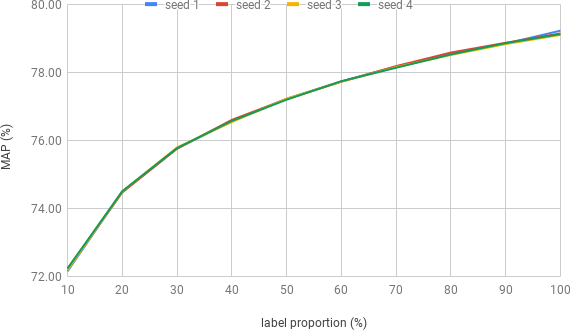} &
\includegraphics[width=1\columnwidth]{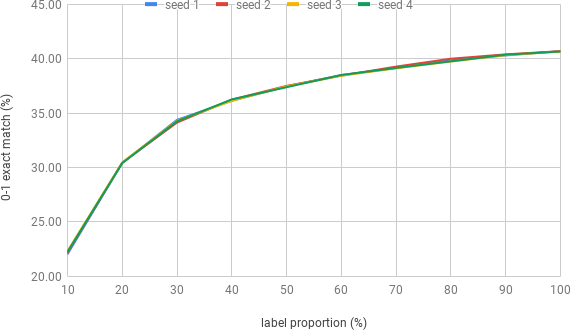}
\\
MAP & 0-1 exact match \\
\includegraphics[width=1\columnwidth]{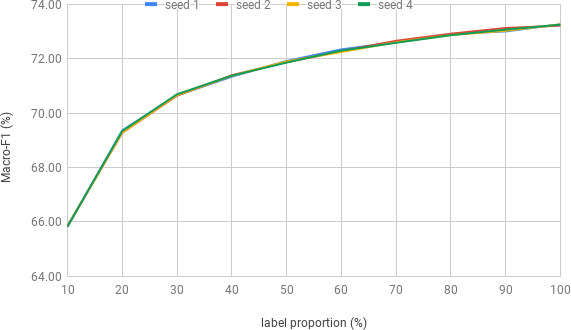} &
\includegraphics[width=1\columnwidth]{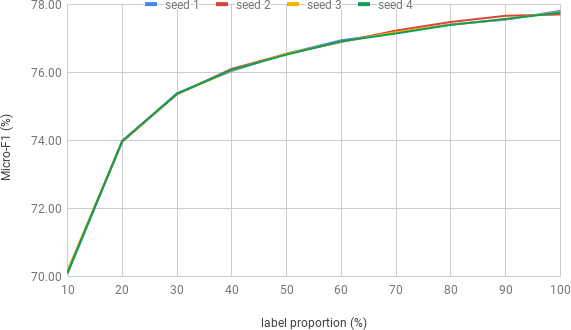}
\\
Macro-F1 & Micro-F1 \\
\includegraphics[width=1\columnwidth]{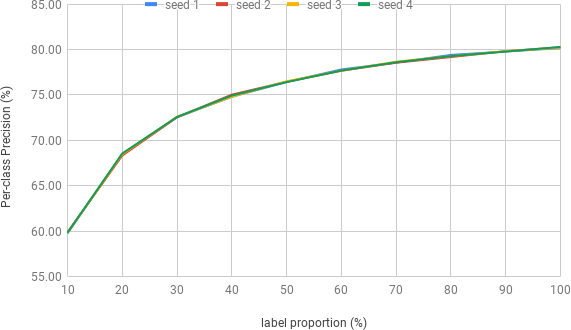} &
\includegraphics[width=1\columnwidth]{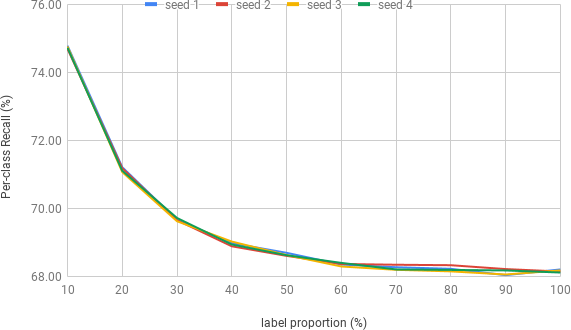}
\\
Per-class Precision & Per-class Recall \\
\includegraphics[width=1\columnwidth]{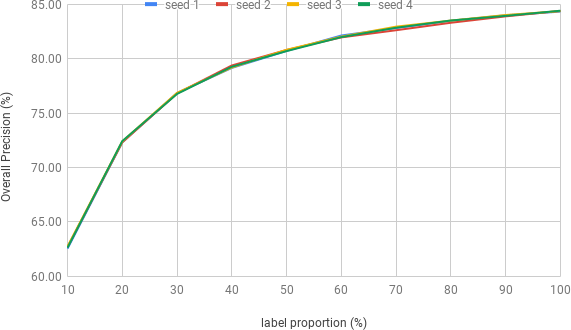} &
\includegraphics[width=1\columnwidth]{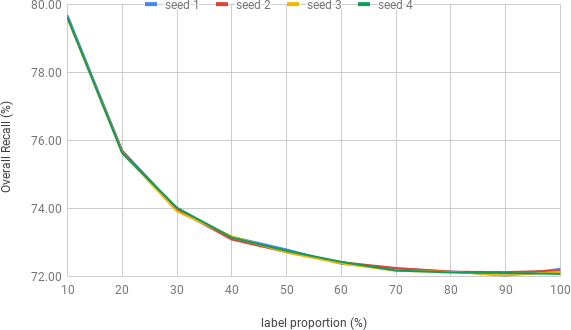}
\\
Overall Precision & Overall Recall \\
\end{tabular}
\caption{Results for differents metrics on MS COCO val2014 to analyze the sensibility of the initial label set.}
\label{fig:seed_analysis}
\end{figure*}

In this section, we analyse the initial set of labels for the partial label scenario.
We report the results for 4 random seeds to generate the initial set of partial labels.
The experiments are performed on MS COCO val2014 with a ResNet-101 WELDON.
The results are shown in \autoref{tab:seed_analysis} and \autoref{fig:seed_analysis} for different label proportions and metrics.
For every label proportion and every metric, we observe that the model is robust to the initial set of labels.

\begin{sidewaystable}
\centering
\begin{tabular}{lccccccccccc}
\toprule
\multirow{ 2}{*}{metric} & \multicolumn{10}{c}{label proportion} \\
& 10\% & 20\% & 30\% & 40\% & 50\% & 60\% & 70\% & 80\% & 90\% & 100\% \\
\midrule
MAP & $72.20 \!\pm\! 0.04$ & $74.49 \!\pm\! 0.02$ & $75.77 \!\pm\! 0.02$ & $76.57 \!\pm\! 0.03$ & $77.21 \!\pm\! 0.01$ & $77.73 \!\pm\! 0.01$ & $78.16 \!\pm\! 0.02$ & $78.53 \!\pm\! 0.03$ & $78.85 \!\pm\! 0.02$ & $79.14 \!\pm\! 0.05$ \\
M-F1 & $65.84 \!\pm\! 0.01$ & $69.32 \!\pm\! 0.04$ & $70.66 \!\pm\! 0.02$ & $71.37 \!\pm\! 0.02$ & $71.88 \!\pm\! 0.03$ & $72.29 \!\pm\! 0.04$ & $72.61 \!\pm\! 0.03$ & $72.89 \!\pm\! 0.03$ & $73.05 \!\pm\! 0.06$ & $73.24 \!\pm\! 0.02$ \\
m-F1 & $70.13 \!\pm\! 0.04$ & $73.97 \!\pm\! 0.01$ & $75.36 \!\pm\! 0.01$ & $76.07 \!\pm\! 0.03$ & $76.54 \!\pm\! 0.01$ & $76.91 \!\pm\! 0.02$ & $77.17 \!\pm\! 0.04$ & $77.42 \!\pm\! 0.04$ & $77.58 \!\pm\! 0.05$ & $77.75 \!\pm\! 0.04$ \\
0-1 & $22.21 \!\pm\! 0.12$ & $30.44 \!\pm\! 0.03$ & $34.26 \!\pm\! 0.11$ & $36.18 \!\pm\! 0.07$ & $37.44 \!\pm\! 0.05$ & $38.46 \!\pm\! 0.04$ & $39.16 \!\pm\! 0.07$ & $39.83 \!\pm\! 0.12$ & $40.34 \!\pm\! 0.04$ & $40.67 \!\pm\! 0.02$ \\
PC-P & $59.82 \!\pm\! 0.05$ & $68.45 \!\pm\! 0.10$ & $72.56 \!\pm\! 0.03$ & $74.88 \!\pm\! 0.11$ & $76.45 \!\pm\! 0.04$ & $77.70 \!\pm\! 0.07$ & $78.59 \!\pm\! 0.05$ & $79.28 \!\pm\! 0.10$ & $79.80 \!\pm\! 0.02$ & $80.22 \!\pm\! 0.05$ \\
PC-R & $74.74 \!\pm\! 0.04$ & $71.14 \!\pm\! 0.07$ & $69.66 \!\pm\! 0.04$ & $68.96 \!\pm\! 0.06$ & $68.64 \!\pm\! 0.04$ & $68.35 \!\pm\! 0.04$ & $68.26 \!\pm\! 0.07$ & $68.23 \!\pm\! 0.08$ & $68.12 \!\pm\! 0.09$ & $68.16 \!\pm\! 0.04$ \\
OV-P & $62.66 \!\pm\! 0.09$ & $72.36 \!\pm\! 0.06$ & $76.81 \!\pm\! 0.04$ & $79.24 \!\pm\! 0.10$ & $80.75 \!\pm\! 0.06$ & $82.01 \!\pm\! 0.08$ & $82.79 \!\pm\! 0.14$ & $83.44 \!\pm\! 0.10$ & $83.94 \!\pm\! 0.06$ & $84.36 \!\pm\! 0.04$ \\
OV-R & $79.62 \!\pm\! 0.04$ & $75.66 \!\pm\! 0.04$ & $73.97 \!\pm\! 0.05$ & $73.14 \!\pm\! 0.04$ & $72.74 \!\pm\! 0.04$ & $72.40 \!\pm\! 0.03$ & $72.21 \!\pm\! 0.04$ & $72.14 \!\pm\! 0.01$ & $72.07 \!\pm\! 0.05$ & $72.15 \!\pm\! 0.06$ \\
\bottomrule
\end{tabular}
\caption{Analysis of the initial set of labels for the partial label scenario. The results are averaged for 4 seeds on MS COCO val2014.}
\label{tab:seed_analysis}
\end{sidewaystable}

\clearpage

\subsection{Analysis of the labeling strategies}

\begin{figure*}[t]
\centering
\begin{tabular}{cc}
\includegraphics[width=1\columnwidth]{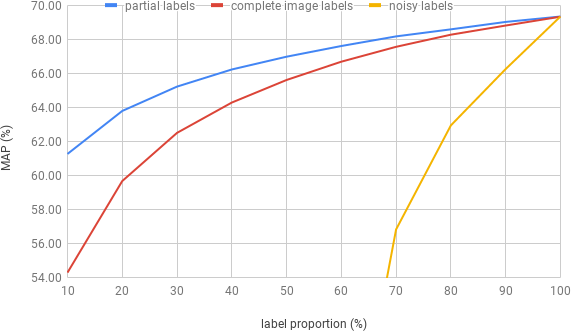} &
\includegraphics[width=1\columnwidth]{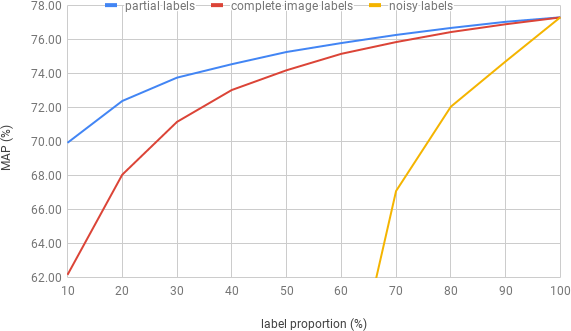}
\\
ResNet-50 & ResNet-50 WELDON \\
\includegraphics[width=1\columnwidth]{image/mscoco/labeling_resnet101weldon_proportion_map} &
\includegraphics[width=1\columnwidth]{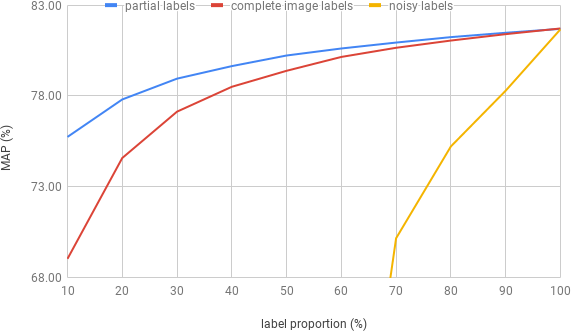}
\\
ResNet-101 WELDON & ResNeXt-101 WELDON
\end{tabular}
\caption{Comparison of the labeling strategies for different label proportions and different architectures on MS COCO val2014.}
\label{fig:app_labeling_archi}
\end{figure*}

In this section we analysis the labeling strategies for different network architectures.
The results are shown in \autoref{tab:app_labeling_archi} and \autoref{fig:app_labeling_archi} on MS COCO dataset.
Overall, the results are very similar.
For a given proportion of labels, we observe that the partial labels strategy is better that the complete image labels.
The improvement increases when the label proportion decreases.
The performance of a model learned with noisy labels drops significantly, even for large proportion of clean labels.

\begin{figure*}[t]
\centering
\begin{tabular}{cc}
\includegraphics[width=1\columnwidth]{image/mscoco/labeling_resnet101weldon_proportion_map} &
\includegraphics[width=1\columnwidth]{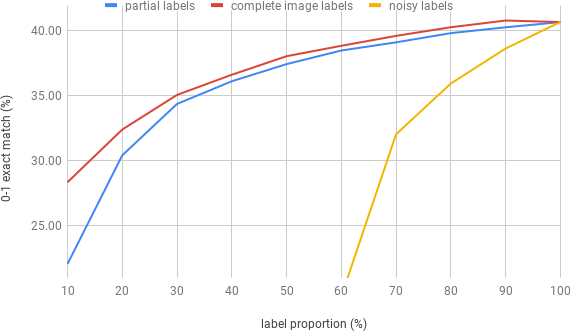}
\\
MAP & 0-1 exact match \\
\includegraphics[width=1\columnwidth]{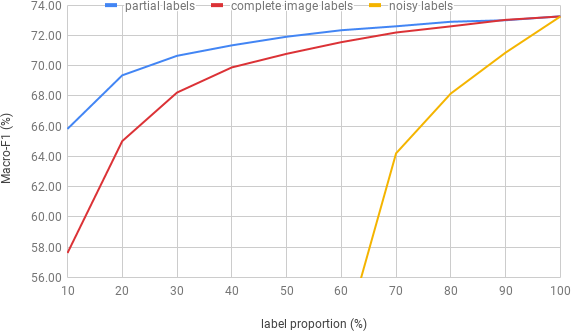} &
\includegraphics[width=1\columnwidth]{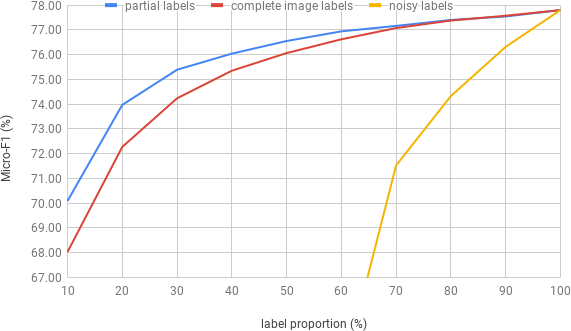}
\\
Macro-F1 & Micro-F1 \\
\includegraphics[width=1\columnwidth]{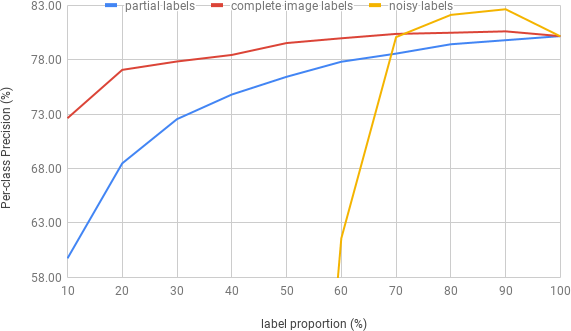} &
\includegraphics[width=1\columnwidth]{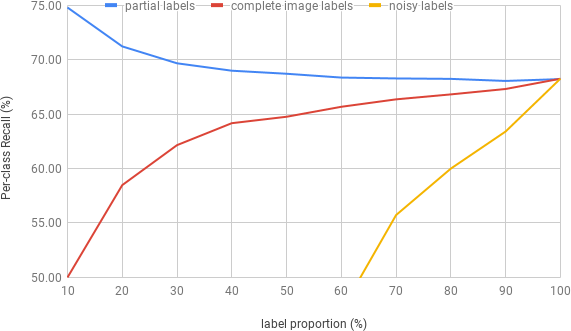}
\\
Per-class Precision & Per-class Recall \\
\includegraphics[width=1\columnwidth]{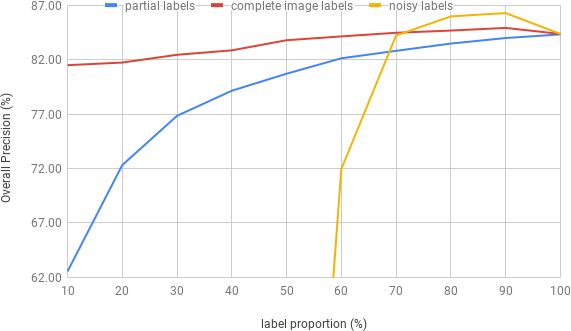} &
\includegraphics[width=1\columnwidth]{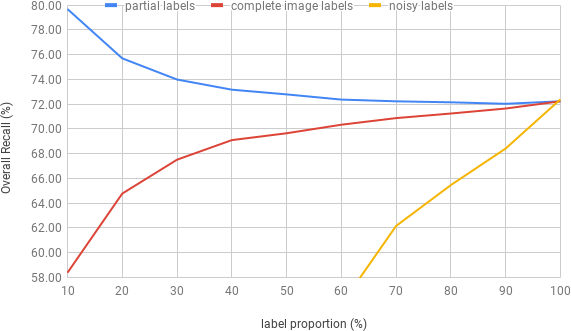}
\\
Overall Precision & Overall Recall \\
\end{tabular}
\caption{Comparison of the labeling strategies for different metrics on MS COCO val2014.}
\label{fig:app_labeling_metrics_mscoco}
\end{figure*}

In \autoref{fig:app_labeling_metrics_mscoco}, we also show the results for different metrics.
For MAP, Macro-F1 and Micro-F1, we observe a similar behaviour: the partial labels strategy has better performances than the complete image labels strategy.
For the 0-1 exact match metric, we observe that the complete image labels strategy has better performances than the complete image labels strategy.
For this metric, the predictions of all the categories must be corrected, so it advantages the complete image labels strategy because some training images have all the labels whereas in the partial labels strategy, none of the training images have all labels.
For the precision and recall metrics, the behaviours are different for the complete image labels strategy and the partial labels strategy.
We note that the complete image labels strategy has a better per-class/overall precision than the partial labels strategy but is has a lower per-class/overall recall than the partial labels strategy.

\begin{table*}
\centering
\begin{tabular}{cccccccccccc}
\toprule
\multirow{ 2}{*}{architecture} & \multirow{ 2}{*}{labels} & \multicolumn{10}{c}{label proportion} \\
& & 10\% & 20\% & 30\% & 40\% & 50\% & 60\% & 70\% & 80\% & 90\% & 100\% \\
\midrule
\multirow{ 3}{*}{ResNet-50} & partial & 61.26 & 63.78 & 65.21 & 66.22 & 66.97 & 67.60 & 68.16 & 68.58 & 69.01 & 69.33 \\
& dense & 54.29 & 59.67 & 62.50 & 64.28 & 65.60 & 66.68 & 67.55 & 68.26 & 68.80 & 69.32 \\
& noisy & - & - & - & - & 3.75 & 39.77 & 56.82 & 62.93 & 66.24 & 69.33 \\
\midrule
\multirow{ 3}{*}{ResNet-50 WELDON} & partial & 69.91 & 72.37 & 73.74 & 74.53 & 75.25 & 75.77 & 76.25 & 76.66 & 77.02 & 77.28 \\
& dense & 62.16 & 68.04 & 71.14 & 73.01 & 74.17 & 75.14 & 75.83 & 76.42 & 76.88 & 77.28 \\
& noisy & - & - & - & - & 3.73 & 52.99 & 67.08 & 72.03 & 74.69 & 77.29 \\
\midrule
\multirow{ 3}{*}{ResNet-101 WELDON} & partial & 72.15 & 74.49 & 75.76 & 76.56 & 77.22 & 77.73 & 78.17 & 78.53 & 78.84 & 79.22 \\
& dense & 65.22 & 71.00 & 73.80 & 75.44 & 76.59 & 77.44 & 78.08 & 78.61 & 78.90 & 79.24 \\
& noisy & - & - & - & - & 3.63 & 53.10 & 69.09 & 74.06 & 76.85 & 79.18 \\
\midrule
\multirow{ 3}{*}{ResNeXt-101 WELDON} & partial & 75.74 & 77.80 & 78.95 & 79.64 & 80.22 & 80.61 & 80.94 & 81.24 & 81.48 & 81.69 \\
& dense & 69.03 & 74.58 & 77.13 & 78.50 & 79.38 & 80.15 & 80.65 & 81.05 & 81.40 & 81.71 \\
& noisy & - & - & - & - & 3.63 & 49.26 & 70.16 & 75.22 & 78.28 & 81.66 \\
\bottomrule
\end{tabular}
\caption{Comparison of the labeling strategies for different label proportions and different architectures on MS COCO val2014.}
\label{tab:app_labeling_archi}
\end{table*}

\paragraph{Comparison to noisy+ strategy.}
In \autoref{tab:app_expes_single_multi}, we show results for the noisy+ strategy on Pascal VOC 2007, MS COCO and NUS-WIDE for different metrics.
For every dataset, we observe that the noisy+ strategy drops the performances of all the metrics with respect to the model learned with only 10\% of clean labels.

\begin{table*}[th]
\centering
\begin{tabular}{llccccccccccc}
\toprule
dataset & strategy & clean label & noisy label & MAP & 0-1 & M-F1 & m-F1 & PC-P & PC-R & OV-P & OV-R \\
\midrule
\multirow{ 3}{*}{VOC 2007} & clean  & 100 & 0 & 93.93 & 79.16 & 88.90 & 91.12 & 90.72 & 87.34 & 93.40 & 88.95 \\
& noisy+        & 97.1 & 2.9 & 90.94 & 62.21 & 78.11 & 78.62 & 95.41 & 68.64 & 97.20 & 66.00 \\
& partial 10\%  & 10 & 0 & 89.09 & 47.46 & 74.55 & 77.84 & 63.35 & 94.16 & 66.02 & 94.81 \\
\midrule
\multirow{ 3}{*}{MS COCO} & clean   & 100 & 0 & 79.22 & 40.69 & 73.26 & 77.80 & 80.16 & 68.21 & 84.31 & 72.23 \\
& noisy+        & 97.6 & 2.4 & 71.60 & 20.28 & 38.62 & 33.72 & 91.76 & 28.17 & 97.34 & 20.39 \\
& partial 10\%  & 10 & 0 & 72.15 & 22.04 & 65.82 & 70.09 & 59.76 & 74.78 & 62.56 & 79.68 \\
\midrule
\multirow{ 3}{*}{NUS-WIDE} & clean         & 100 & 0 & 54.88 & 42.29 & 51.88 & 71.15 & 58.54 & 49.33 & 73.83 & 68.66 \\
& noisy+        & 98.6 & 1.4 & 47.44 & 36.07 & 18.83 & 28.53 & 59.71 & 13.95 & 83.72 & 17.19 \\
& partial 10\%  & 10 & 0 & 51.14 & 25.98 & 51.36 & 65.52 & 41.80 & 69.23 & 53.62 & 84.19 \\
\bottomrule
\end{tabular}
\caption{Comparison with a webly-supervised strategy (noisy+) on MS COCO.
Clean (resp. noisy) means the percentage of clean (resp. noisy) labels in the training set.
Noisy+ is a labeling strategy where there is only one positive label per image.}
\label{tab:app_expes_single_multi}
\end{table*}

\clearpage

\subsection{Comparison of the loss functions}
\label{sec:app_compa_loss}

\begin{figure*}[t]
\centering
\begin{tabular}{cc}
\includegraphics[width=1\columnwidth]{image/mscoco/resnet101weldon_proportion_map} &
\includegraphics[width=1\columnwidth]{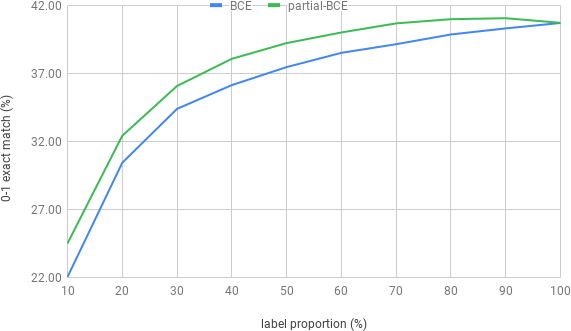}
\\
MAP & 0-1 exact match \\
\includegraphics[width=1\columnwidth]{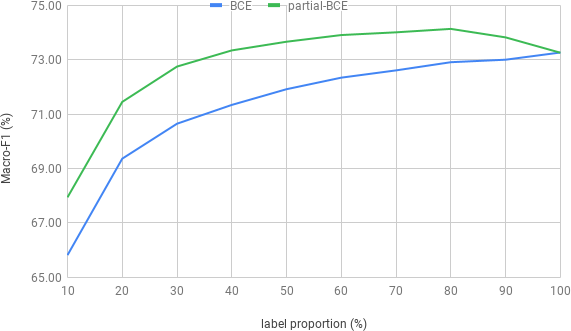} &
\includegraphics[width=1\columnwidth]{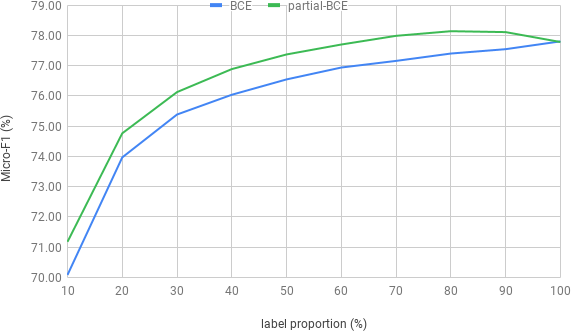}
\\
Macro-F1 & Micro-F1 \\
\includegraphics[width=1\columnwidth]{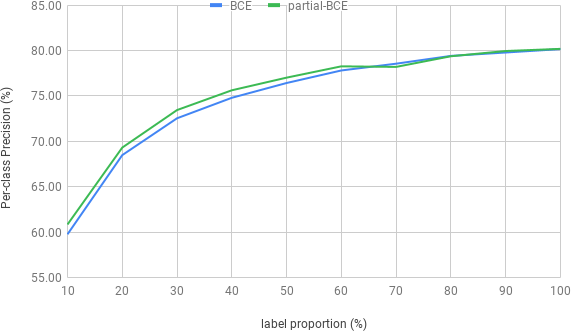} &
\includegraphics[width=1\columnwidth]{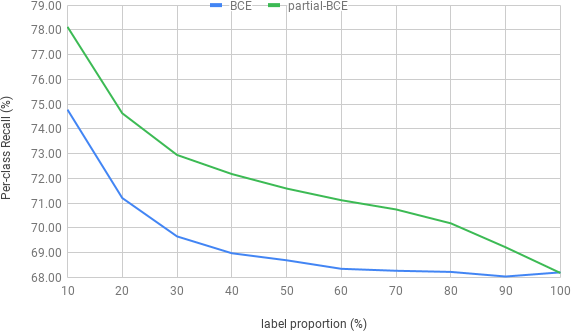}
\\
Per-class Precision & Per-class Recall \\
\includegraphics[width=1\columnwidth]{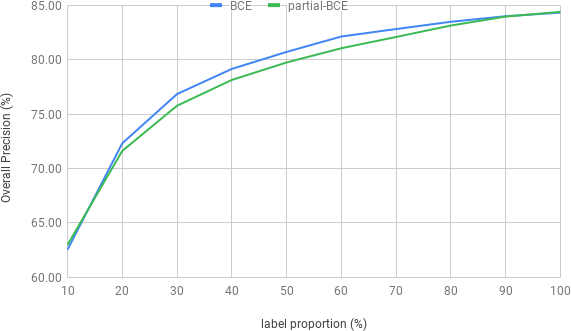} &
\includegraphics[width=1\columnwidth]{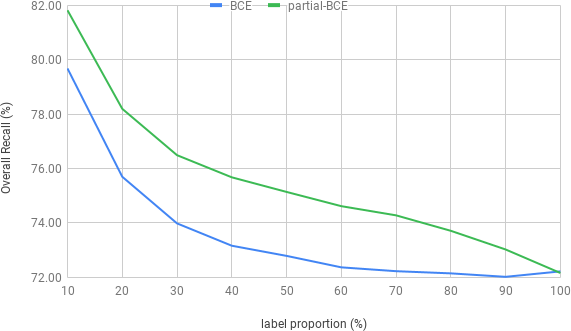}
\\
Overall Precision & Overall Recall \\
\end{tabular}
\caption{Results for different metrics on MS COCO val2014.}
\label{fig:app_proportion_metrics_mscoco}
\end{figure*}

\begin{figure*}[t]
\centering
\begin{tabular}{cc}
\includegraphics[width=1\columnwidth]{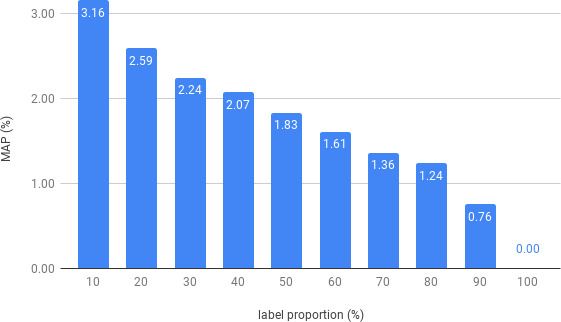} &
\includegraphics[width=1\columnwidth]{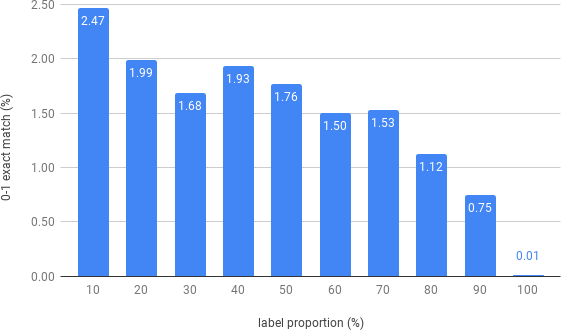}
\\
MAP & 0-1 exact match \\
\includegraphics[width=1\columnwidth]{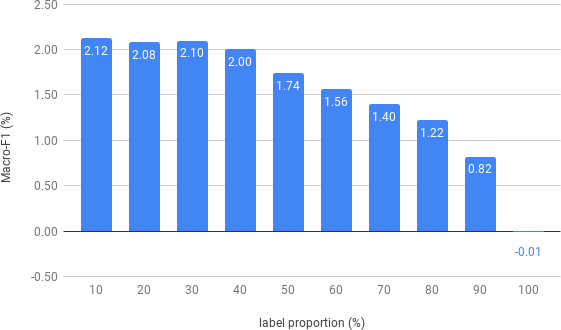} &
\includegraphics[width=1\columnwidth]{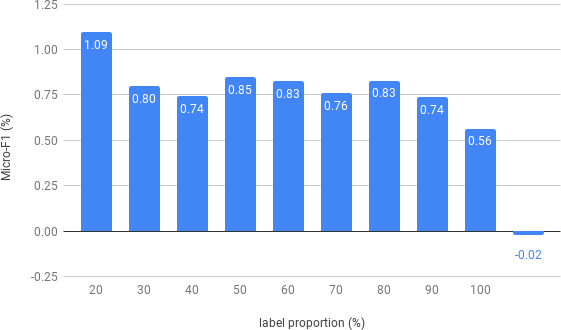}
\\
Macro-F1 & Micro-F1 \\
\includegraphics[width=1\columnwidth]{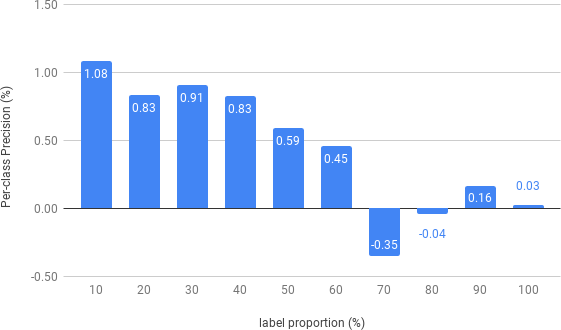} &
\includegraphics[width=1\columnwidth]{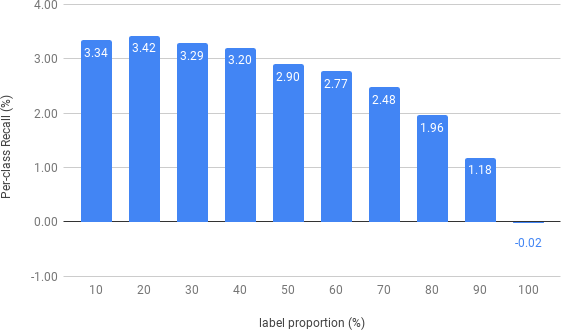}
\\
Per-class Precision & Per-class Recall \\
\includegraphics[width=1\columnwidth]{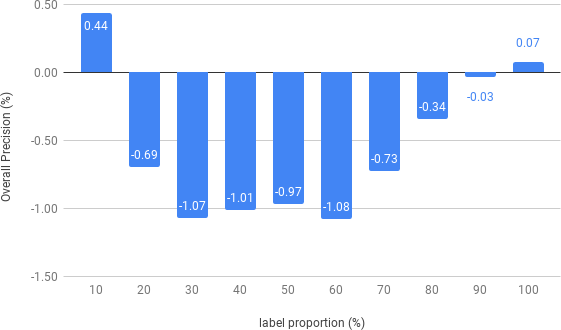} &
\includegraphics[width=1\columnwidth]{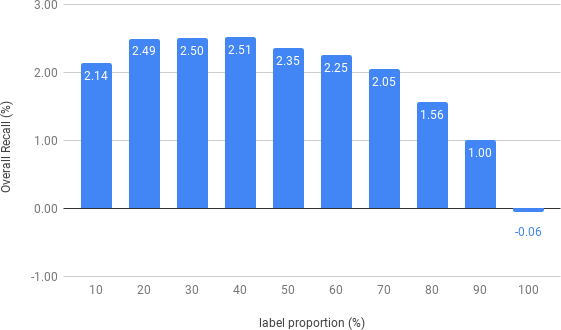}
\\
Overall Precision & Overall Recall \\
\end{tabular}
\caption{Improvement analysis between partial-BCE and BCE for differents metrics on MS COCO val2014.}
\label{fig:app_proportion_metrics_mscoco_diff}
\end{figure*}

\begin{figure*}[p]
\centering
\begin{tabular}{cc}
\includegraphics[width=1\columnwidth]{image/voc2007/resnet101weldon_proportion_map} &
\includegraphics[width=1\columnwidth]{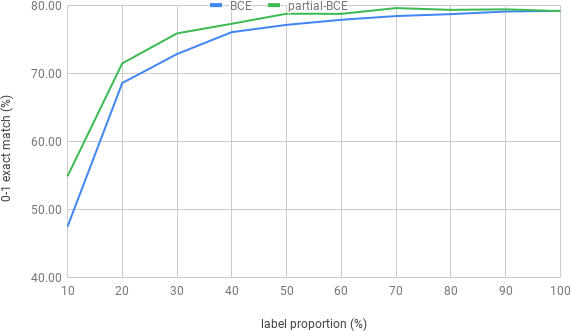}
\\
MAP & 0-1 exact match \\
\includegraphics[width=1\columnwidth]{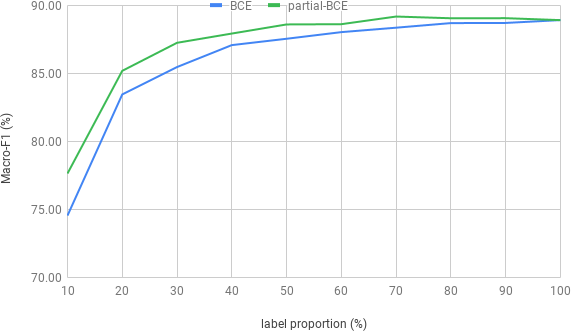} &
\includegraphics[width=1\columnwidth]{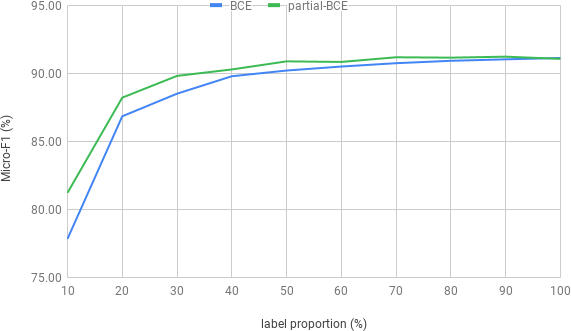}
\\
Macro-F1 & Micro-F1 \\
\includegraphics[width=1\columnwidth]{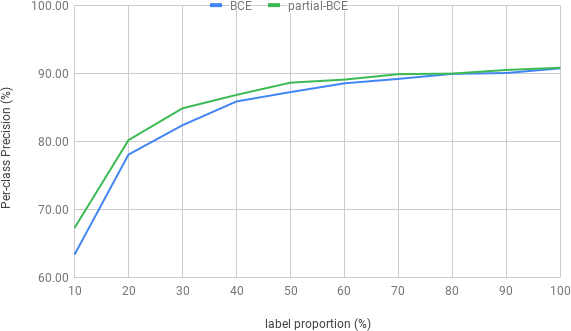} &
\includegraphics[width=1\columnwidth]{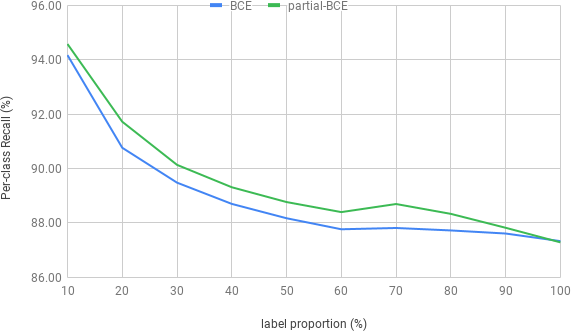}
\\
Per-class Precision & Per-class Recall \\
\includegraphics[width=1\columnwidth]{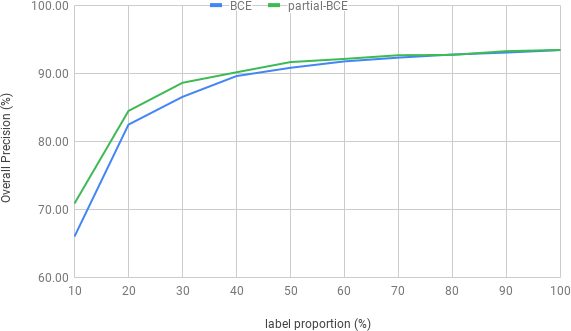} &
\includegraphics[width=1\columnwidth]{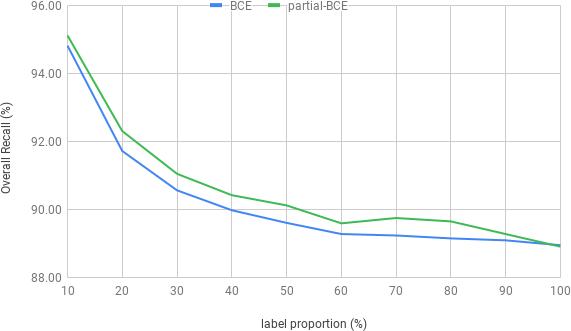}
\\
Overall Precision & Overall Recall \\
\end{tabular}
\caption{Results for different metrics on Pascal VOC 2007.}
\label{fig:app_proportion_metrics_voc2007}
\end{figure*}

\begin{figure*}[p]
\centering
\begin{tabular}{cc}
\includegraphics[width=1\columnwidth]{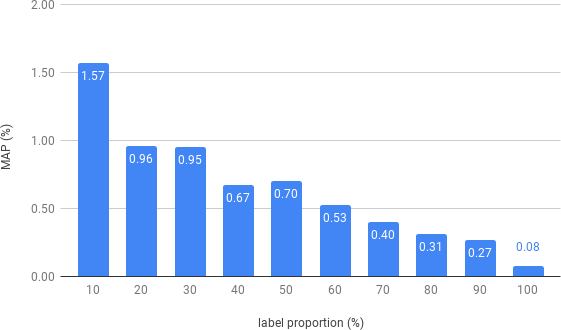} &
\includegraphics[width=1\columnwidth]{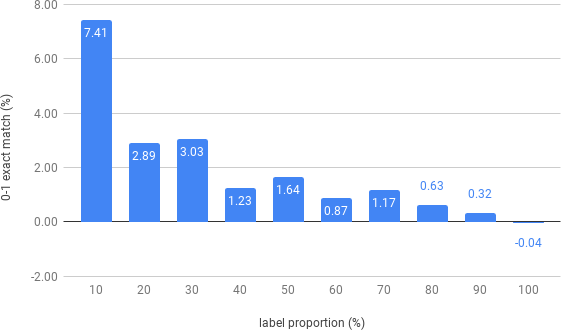}
\\
MAP & 0-1 exact match \\
\includegraphics[width=1\columnwidth]{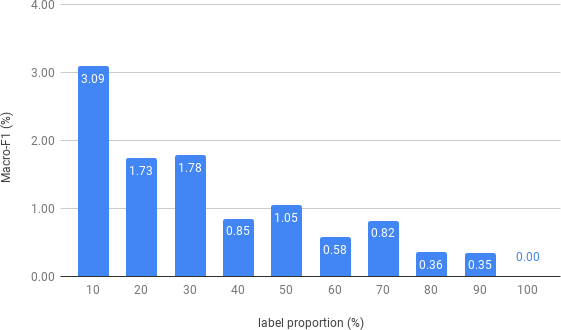} &
\includegraphics[width=1\columnwidth]{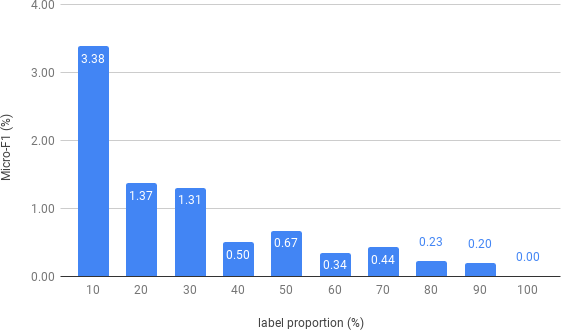}
\\
Macro-F1 & Micro-F1 \\
\includegraphics[width=1\columnwidth]{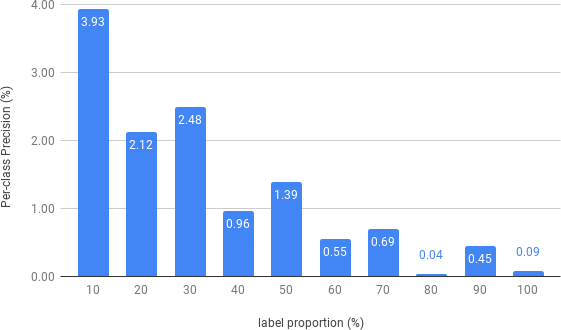} &
\includegraphics[width=1\columnwidth]{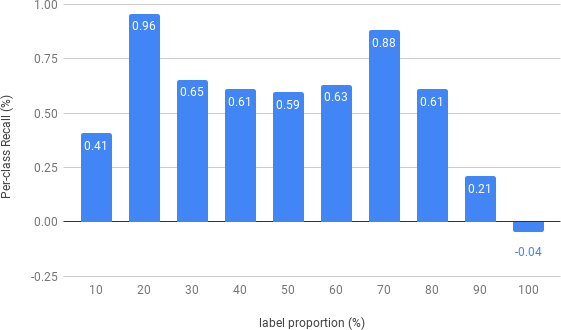}
\\
Per-class Precision & Per-class Recall \\
\includegraphics[width=1\columnwidth]{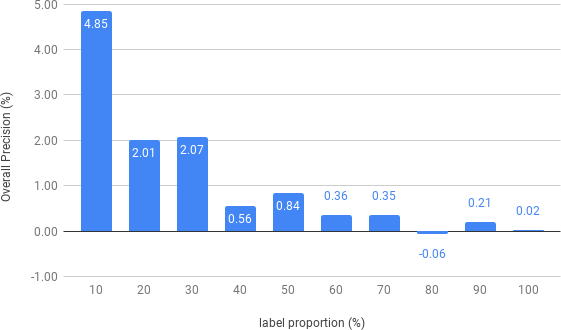} &
\includegraphics[width=1\columnwidth]{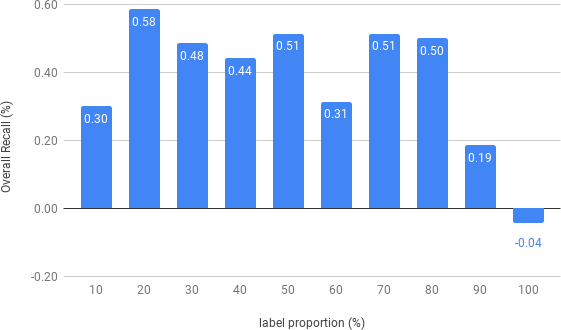}
\\
Overall Precision & Overall Recall \\
\end{tabular}
\caption{Improvement analysis between partial-BCE and BCE for differents metrics on Pascal VOC 2007.}
\label{fig:app_proportion_metrics_voc2007_diff}
\end{figure*}

In this section, we analyse the performances of the BCE and partial-BCE loss functions for different metrics.
The results on MS COCO (resp. Pascal VOC 2007) are shown in \autoref{fig:app_proportion_metrics_mscoco} (resp. \autoref{fig:app_proportion_metrics_voc2007}) and the improvement of the partial-BCE with respect to the BCE is shown in \autoref{fig:app_proportion_metrics_mscoco_diff} (resp. \autoref{fig:app_proportion_metrics_voc2007_diff}).
We observe that the partial-BCE significantly improves the performances for MAP, 0-1 exact match, Macro-F1 and Micro-F1 metrics.
We note that the improvement is bigger when the label proportion is lower.
The proposed loss also improves the (overall and per-class) recall for both datasets.
On Pascal VOC 2007, it also improves the overall and per-class precision.
However, we observe that the

\begin{figure}[h]
\centering
\includegraphics[width=0.95\columnwidth]{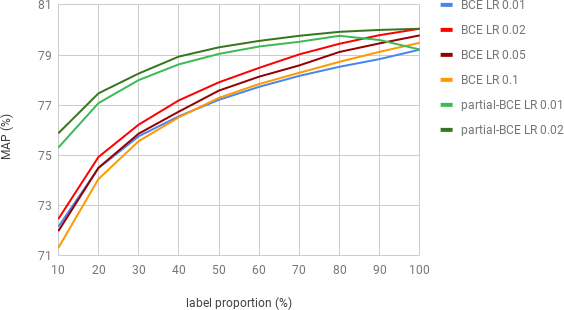}
\caption{Analysis of the learning rate on MS COCO dataset.}
\label{fig:lr_coco}
\vspace{-3mm}
\end{figure}

We observe that decreasing the proportion of known labels can slightly improves the performances with respect to the model trained with all the annotations.
This phenomenon is because of the tuning of the learning rate and the hyperparameter $\gamma$ (\autoref{fig:expes_weight_gamma}).
Note that the BCE and the partial-BCE have the same results for the label proportion 100\% because they are equivalent by definition.
In the paper, we used the same training setting (learning rate, weight decay, \etc) as \cite{Durand2018} for each model and dataset.
In \autoref{fig:lr_coco}, we observe that using a learning rate of 0.02 increases the performance and leads to a monotone increase of the performance with respect to the label proportion, but the optimal learning rate depends on the dataset.
It is possible to improve the results by tuning carefully these hyperparameters, but we observe that the partial-BCE is still better than the BCE for a large range of LRs and for small label proportions which is the main focus of the paper.

\clearpage

\subsection{Analysis of the loss function}
\label{sec:app_loss_analysis}

\begin{figure*}[ht!]
\centering
\begin{tabular}{cc}
\includegraphics[width=1\columnwidth]{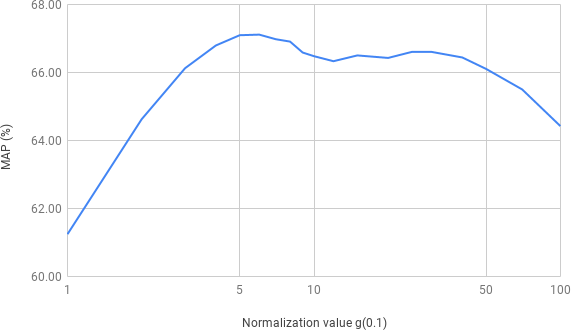} &
\includegraphics[width=1\columnwidth]{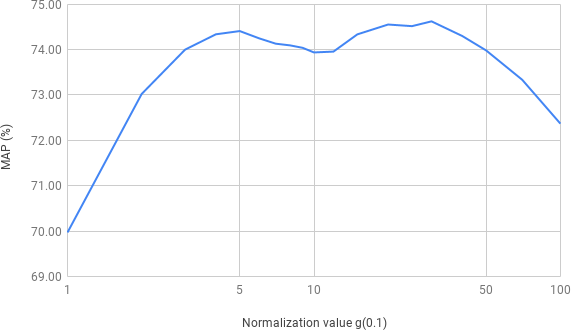}
\\
ResNet-50 & ResNet-50 WELDON \\
\includegraphics[width=1\columnwidth]{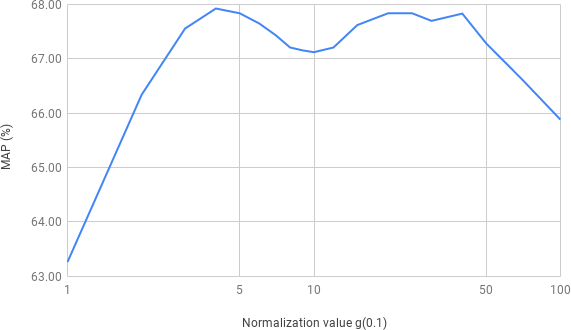} &
\includegraphics[width=1\columnwidth]{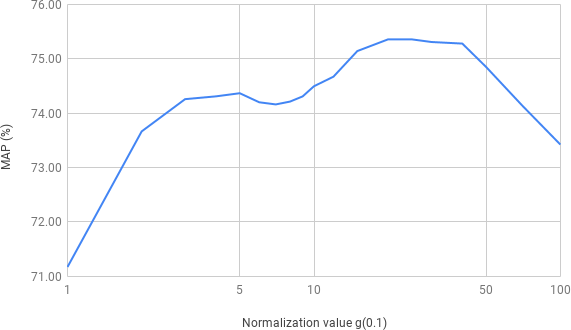}
\\
ResNet-101 & ResNet-101 WELDON
\end{tabular}
\caption{Analysis of the normalization value for 10\% of known labels (\ie $g(0.1)$) on MS COCO. (x-axis log-scale)}
\label{fig:app_analysis_g}
\end{figure*}

In this section, we analyze the hyperparameter of the loss function for several network architectures.
The models are trained on the train2014 set minus 5000 images that are used as validation set to evaluate the performances.
The \autoref{fig:app_analysis_g} shows the results on MS COCO.
We observe a similar behavior for all the architectures.
Overall, using a normalization value $g(0.1)$ between 3 and 50 significantly improves the performances with respect to the normalization by the number of categories ($g(0.1)=1$).
The loss is robust to the value of this hyperparmeter.

\clearpage

\subsection{Comparison to existing model for missing labels}

As pointed out in the related work section, most of the existing models to learn with missing labels are not scalable and do not allow experiments on large-scale dataset like MS COCO and NUS-WIDE.
We compare our model with the APG-Graph model \cite{Yang2016} that models structured semantic correlations between images on the Pascal VOC 2007 dataset.
Unlike our method, the APG-Graph model does not allow to fine-tune the ConvNet.

\begin{figure}[h]
\centering
\includegraphics[width=1\columnwidth]{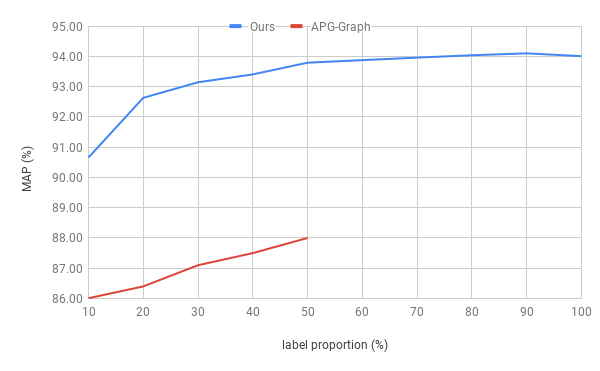}
\caption{Comparison with APG-Graph model on Pascal VOC 2007 for different proportion of known labels. }
\label{fig:sota_missing_label}
\end{figure}

\subsection{What is the best strategy to predict missing labels?}
\label{sec:app_label_missing}

This section extends the section 4.3 in the paper.
First, to compute the Bayesian uncertainty, we use the setting used in the original paper \cite{Kendall2017}.
The results for different strategies and hyperparameters are shown in \autoref{tab:app_expes_relabeling_voc2007}.
G defines how the examples are selected during training.
In the paper, we only explain how to find the solution with respect to $\mathbf{v}$.
G depends on the strategy and is defined as:
\begin{align*}
G(\mathbf{v}; \theta) = - \sum_{i=1}^N \sum_{c=1}^C  v_{ic} \log \left( \frac{1}{1 + e^{-\theta}} \right)
\end{align*}
for strategy [a].

For strategy [a] and [d], we observe that using a small threshold is better than a large threshold.
On the contrary, for strategy [c] we observe that using a large threshold is better than a small threshold, but the results are worse than strategy [a].
For strategy [b], labeling a large proportion of labels per mini-batch is better than labeling a small proportion of labels.
For strategy [e], we note that using a GNN improves the performances of the model and the model is more robust to the threshold hyperparameter $\theta$.

\begin{table*}[t]
\centering
\begin{tabular}{lcccc|cccc}
\toprule
Relabeling & MAP & 0-1 & Macro-F1 & Micro-F1 & label prop. & TP & TN & GNN \\
\midrule
2 steps (no curriculum) & -1.49 & 6.42 & 2.32 & 1.99 & 100 & 82.78 & 96.40 & \cmark \\
\midrule
\textbf{[a]} Score threshold $\theta=1$ & 0.00 & 11.31 & 3.71 & 4.25 & 97.87 & 82.47 & 97.84 & \cmark \\
\textbf{[a]} Score threshold $\theta=2$ & 0.34 & 11.15 & 4.33 & 4.26 & 95.29 & 85.00 & 98.50 & \cmark \\
\textbf{[a]} Score threshold $\theta=5$ & 0.31 & 5.02 & 2.60 & 1.83 & 70.98 & 96.56 & 99.44 & \cmark \\
\midrule
\textbf{[b]} Score proportion $\theta=0.1$ & 0.45 & -1.20 & -0.28 & -0.68 & 26.70 & 99.28 & 99.19 & \cmark \\
\textbf{[b]} Score proportion $\theta=0.2$ & 0.36 & 0.20 & 0.70 & 0.10 & 42.09 & 98.35 & 99.33 & \cmark \\
\textbf{[b]} Score proportion $\theta=0.3$ & 0.28 & 0.91 & 1.09 & 0.37 & 55.63 & 97.82 & 99.38 & \cmark \\
\textbf{[b]} Score proportion $\theta=0.4$ & 0.55 & 2.95 & 2.33 & 1.28 & 67.41 & 96.87 & 99.38 & \cmark \\
\textbf{[b]} Score proportion $\theta=0.5$ & 0.22 & 4.02 & 2.76 & 1.74 & 77.40 & 95.52 & 99.30 & \cmark \\
\textbf{[b]} Score proportion $\theta=0.6$ & 0.41 & 6.17 & 3.63 & 2.52 & 85.37 & 93.16 & 99.15 & \cmark \\
\textbf{[b]} Score proportion $\theta=0.7$ & 0.35 & 7.49 & 3.83 & 3.07 & 91.69 & 89.40 & 98.81 & \cmark \\
\textbf{[b]} Score proportion $\theta=0.8$ & 0.17 & 8.40 & 3.70 & 3.25 & 96.24 & 84.40 & 98.10 & \cmark \\
\midrule
\textbf{[c]} Postitive only - score $\theta=1$ & -1.61 & -31.75 & -18.07 & -18.92 & 16.79 & 36.42 & - & \cmark \\
\textbf{[c]} Postitive only - score $\theta=2$ & -0.80 & -21.31 & -10.93 & -12.08 & 14.71 & 47.94 & - & \cmark \\
\textbf{[c]} Postitive only - score $\theta=5$ & 0.31 & -4.58 & -1.92 & -2.23 & 12.01 & 79.07 & - & \cmark \\
\midrule
\textbf{[d]} Ensemble score $\theta=1$ & -0.31 & 10.16 & 3.61 & 3.94 & 97.84 & 82.12 & 97.76 & \cmark \\
\textbf{[d]} Ensemble score $\theta=2$ & 0.23 & 11.31 & 4.16 & 4.33 & 95.33 & 84.80 & 98.53 & \cmark \\
\textbf{[d]} Ensemble score $\theta=5$ & 0.27 & 3.78 & 2.38 & 1.53 & 70.77 & 96.56 & 99.44 & \cmark \\
\midrule
\midrule
\textbf{[e]} Bayesian uncertainty $\theta=0.1$ & 0.26 & 1.84 & 1.36 & 0.64 & 22.63 & 25.71 & 99.98 & \\
\textbf{[e]} Bayesian uncertainty $\theta=0.2$ & 0.29 & 8.49 & 4.05 & 3.66 & 60.32 & 48.39 & 99.82 & \\
\textbf{[e]} Bayesian uncertainty $\theta=0.3$ & 0.34 & 10.15 & 4.37 & 3.72 & 77.91 & 61.15 & 99.24 & \\
\textbf{[e]} Bayesian uncertainty $\theta=0.4$ & 0.30 & 9.05 & 4.17 & 3.37 & 87.80 & 68.56 & 98.70 & \\
\textbf{[e]} Bayesian uncertainty $\theta=0.5$ & 0.26 & 8.32 & 3.83 & 3.05 & 92.90 & 70.96 & 98.04 & \\
\midrule
\textbf{[e]} Bayesian uncertainty $\theta=0.1$ & 0.36 & 2.71 & 1.91 & 1.22 & 19.45 & 38.15 & 99.97 & \cmark \\
\textbf{[e]} Bayesian uncertainty $\theta=0.2$ & 0.30 & 10.76 & 4.87 & 4.66 & 57.03 & 62.03 & 99.65 & \cmark \\
\textbf{[e]} Bayesian uncertainty $\theta=0.3$ & 0.59 & 12.07 & 5.11 & 4.95 & 79.74 & 68.96 & 99.23 & \cmark \\
\textbf{[e]} Bayesian uncertainty $\theta=0.4$ & 0.43 & 10.99 & 4.88 & 4.46 & 90.51 & 70.77 & 98.57 & \cmark \\
\textbf{[e]} Bayesian uncertainty $\theta=0.5$ & 0.45 & 10.08 & 3.93 & 3.78 & 94.79 & 74.73 & 98.00 & \cmark \\
\bottomrule
\end{tabular}
\caption{Analysis of the labeling strategy of missing labels on Pascal VOC 2007 val set.
For each metric, we report the relative scores with respect to a model that does not label missing labels.
TP (resp. TN) means true positive (resp. true negative).
Label proportion is the proportion of training labels (clean + weak labels) used at the end of the training.
For the strategy labeling only positive labels, we report the label accuracy instead of the TP rate.
}
\label{tab:app_expes_relabeling_voc2007}
\end{table*}

\clearpage
\subsection{Final results}

\begin{figure*}[t]
\centering
\begin{tabular}{cc}
\includegraphics[width=1\columnwidth]{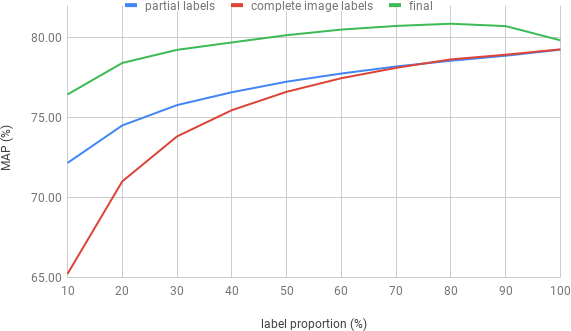} &
\includegraphics[width=1\columnwidth]{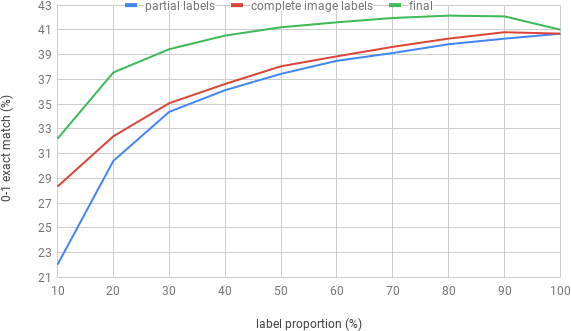}
\\
MAP & 0-1 exact match \\
\includegraphics[width=1\columnwidth]{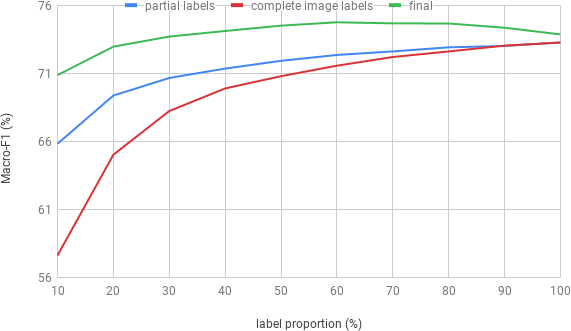} &
\includegraphics[width=1\columnwidth]{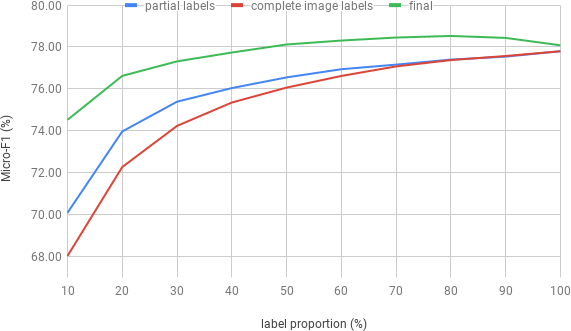}
\\
Macro-F1 & Micro-F1 \\
\includegraphics[width=1\columnwidth]{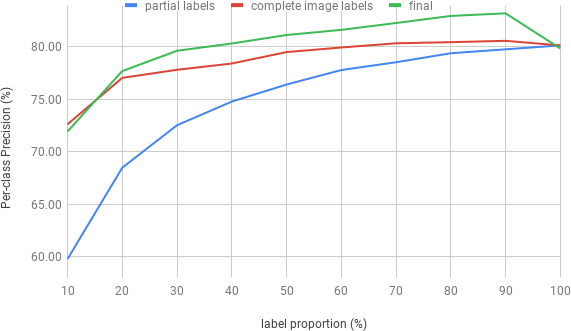} &
\includegraphics[width=1\columnwidth]{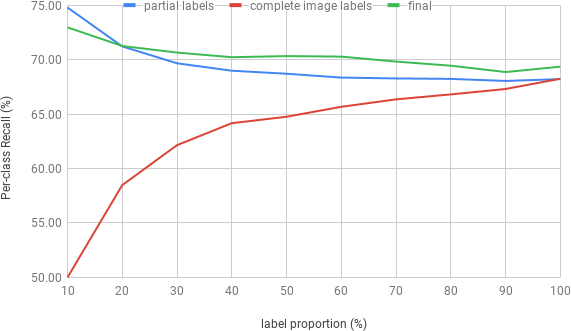}
\\
Per-class Precision & Per-class Recall \\
\includegraphics[width=1\columnwidth]{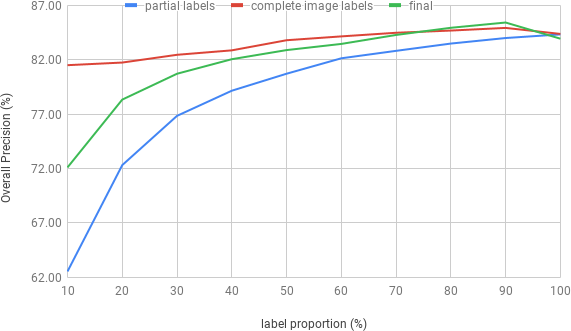} &
\includegraphics[width=1\columnwidth]{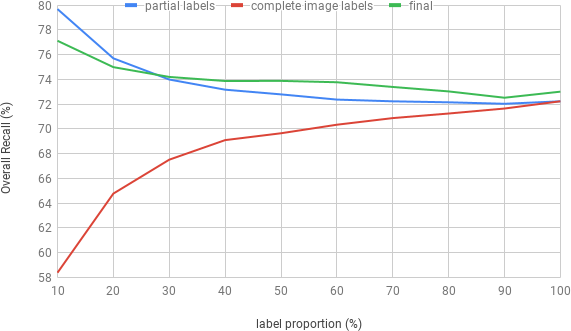}
\\
Overall Precision & Overall Recall \\
\end{tabular}
\caption{The results of our final model with two baselines (complete image labeling and BCE with partial labels) for different metrics on MS COCO val2014.}
\label{fig:app_labeling_metrics_mscoco_final}
\end{figure*}

In \autoref{fig:app_labeling_metrics_mscoco_final}, we show the results of our final model that uses the partial-BCE loss, the GNN and the labeling of missing labels.
We compare our model to two baselines: (a) a model trained with the standard BCE where the data are labeled with the partial labels strategy (blue) and (b) a model trained with the standard BCE where the data are labeled with the complete image labels strategy (red).
We observe that our model has better performances than the two baselines for most of the metrics.
In particular, our final model has significantly better 0-1 exact match performance than the baseline (b), whereas the baseline with partial labels (a) has lower performance than the baseline (b).
We note that the overall precision of our model is worse than the baseline (b), but the overall recall of our model is largely better than the baseline (b).

%% file: appendix_gnn.tex
\subsection{Multi-label classification with GNN}
\label{sec:app_gnn}

In this section, we give additional information about the Graph Neural Networks (GNN) used in our work.
We first show the algorithm used to predict the classification scores with a GNN in Algorithm \autoref{alg:app_gnn}.
The input $\x \in \mathbb R^C$ of the GNN is the ConvNet output, where $C$ is the number of categories.

The $f_\mathcal{M}$ function in the message update function $\mathcal{M}$ is a fully connected layer followed by a ReLU.
Because the graph is fully-connected, the message update function $\mathcal{M}$ averages on all the nodes of the graph excepts the current node $v$ \ie $\Omega_v = \cv \setminus \{v\}$.
Similarly to \cite{Qi2017}, the final prediction uses both first and last hidden states.
We observe that using both first and last hidden states is better than using only the last hidden state.
According to \cite{Qi2017}, we use $T=3$ iterations in our experiments.

\begin{algorithm}[h]
\caption{Graph Neural Network (GNN)}
\label{alg:app_gnn}
\begin{algorithmic}[1]
\REQUIRE ConvNet output $\x$
\STATE Initialize the hidden state of each node $v \in \cv$ with the output of the ConvNet.
\begin{equation}
\h_v^0 = [0, \ldots, 0, x_v, 0, \ldots, 0] \qquad \forall v \in \cv
\end{equation}
\FOR{t = 0 \TO T-1 }
\STATE Update message of each node $v \in \cv$ based on the hidden states
\begin{equation}
\m_v^t = \mathcal{M}(\{\h_u^t | u \in \Omega_v\}) = \frac{1}{|\Omega_v| } \sum_{u \in \Omega_v} f_\mathcal{M}(\h_u^t)
\end{equation}
\STATE Update hidden state of each node $v \in \cv$ based on the messages
\begin{equation}
\h_v^{t+1} = \mathcal{F}(\h_v^t, \m_v^t) = GRU(\h_v^t, \m_v^t)
\end{equation}
\ENDFOR
\STATE Compute the output based on the first and last hidden states
\begin{equation}
\bar{\y} = s(\h_v^0, \h_v^T) = \h_v^0 + \h_v^T
\end{equation}
\ENSURE $\bar{\y}$
\end{algorithmic}
\end{algorithm}

%% file: main_arxiv.bbl
\begin{thebibliography}{10}\itemsep=-1pt

\bibitem{BaezaYates1999}
R.~A. Baeza-Yates and B.~Ribeiro-Neto.
\newblock {\em Modern Information Retrieval}.
\newblock 1999.

\bibitem{Bengio2009}
Y.~Bengio, J.~Louradour, R.~Collobert, and J.~Weston.
\newblock Curriculum learning.
\newblock In {\em International Conference on Machine Learning (ICML)}, 2009.

\bibitem{Bucak2011}
S.~S. Bucak, R.~Jin, and A.~K. Jain.
\newblock {Multi-label learning with incomplete class assignments}.
\newblock In {\em {IEEE} Conference on Computer Vision and Pattern Recognition
  (CVPR)}, 2011.

\bibitem{Cabral2011}
R.~S. Cabral, F.~Torre, J.~P. Costeira, and A.~Bernardino.
\newblock {Matrix Completion for Multi-label Image Classification}.
\newblock In {\em Advances in Neural Information Processing Systems (NIPS)},
  2011.

\bibitem{Carlson2010}
A.~Carlson, J.~Betteridge, B.~Kisiel, B.~Settles, E.~R.~H. Jr., and T.~M.
  Mitchell.
\newblock {Toward an Architecture for Never-Ending Language Learning}.
\newblock In {\em Conference on Artificial Intelligence (AAAI)}, 2010.

\bibitem{Chapelle2010}
O.~Chapelle, B.~Schlkopf, and A.~Zien.
\newblock {\em Semi-Supervised Learning}.
\newblock 2010.

\bibitem{Chen2013}
X.~Chen, A.~Shrivastava, and A.~Gupta.
\newblock Neil: Extracting visual knowledge from web data.
\newblock In {\em {IEEE} International Conference on Computer Vision (ICCV)},
  2013.

\bibitem{Chen2014}
X.~Chen, A.~Shrivastava, and A.~Gupta.
\newblock Enriching visual knowledge bases via object discovery and
  segmentation.
\newblock In {\em {IEEE} Conference on Computer Vision and Pattern Recognition
  (CVPR)}, 2014.

\bibitem{Cho2014}
K.~Cho, B.~{van Merrienboer}, D.~Bahdanau, and Y.~Bengio.
\newblock {On the properties of neural machine translation: Encoder-decoder
  approaches}.
\newblock In {\em Eighth Workshop on Syntax, Semantics and Structure in
  Statistical Translation (SSST-8)}, 2014.

\bibitem{Chu2018}
H.-M. Chu, C.-K. Yeh, and Y.-C. Frank~Wang.
\newblock {Deep Generative Models for Weakly-Supervised Multi-Label
  Classification}.
\newblock In {\em European Conference on Computer Vision ({ECCV})}, 2018.

\bibitem{Chua2009}
T.-S. Chua, J.~Tang, R.~Hong, H.~Li, Z.~Luo, and Y.~Zheng.
\newblock {NUS-WIDE: A Real-world Web Image Database from National University
  of Singapore}.
\newblock In {\em ACM International Conference on Image and Video Retrieval
  (CIVR)}, 2009.

\bibitem{Cour2011}
T.~Cour, B.~Sapp, and B.~Taskar.
\newblock {Learning from Partial Labels}.
\newblock {\em Journal of Machine Learning Research (JMLR)}, 2011.

\bibitem{Deng2014b}
J.~Deng, O.~Russakovsky, J.~Krause, M.~S. Bernstein, A.~Berg, and L.~Fei-Fei.
\newblock {Scalable Multi-label Annotation}.
\newblock In {\em Proceedings of the SIGCHI Conference on Human Factors in
  Computing Systems}, 2014.

\bibitem{Durand2017}
T.~Durand, T.~Mordan, N.~Thome, and M.~Cord.
\newblock {WILDCAT: Weakly Supervised Learning of Deep ConvNets for Image
  Classification, Pointwise Localization and Segmentation}.
\newblock In {\em {IEEE} Conference on Computer Vision and Pattern Recognition
  (CVPR)}, 2017.

\bibitem{Durand2016}
T.~Durand, N.~Thome, and M.~Cord.
\newblock {WELDON: Weakly Supervised Learning of Deep Convolutional Neural
  Networks}.
\newblock In {\em {IEEE} Conference on Computer Vision and Pattern Recognition
  (CVPR)}, 2016.

\bibitem{Durand2018}
T.~Durand, N.~Thome, and M.~Cord.
\newblock {Exploiting Negative Evidence for Deep Latent Structured Models}.
\newblock In {\em {IEEE} Transactions on Pattern Analysis and Machine
  Intelligence (TPAMI)}, 2018.

\bibitem{Everingham2015}
M.~Everingham, S.~M.~A. Eslami, L.~Van~Gool, C.~K.~I. Williams, J.~Winn, and
  A.~Zisserman.
\newblock {The Pascal Visual Object Classes Challenge: A Retrospective}.
\newblock {\em International Journal of Computer Vision (IJCV)}, 2015.

\bibitem{Gong2014}
Y.~Gong, Y.~Jia, T.~Leung, A.~Toshev, and S.~Ioffe.
\newblock {Deep Convolutional Ranking for Multilabel Image Annotation}.
\newblock In {\em International Conference on Learning Representations (ICLR)},
  2014.

\bibitem{Gori2005}
M.~Gori, G.~Monfardini, and F.~Scarselli.
\newblock {A new model for learning in graph domains}.
\newblock In {\em {IEEE} International Joint Conference on Neural Networks
  (IJCNN)}, 2005.

\bibitem{Guo2018}
S.~Guo, W.~Huang, H.~Zhang, C.~Zhuang, D.~Dong, M.~R. Scott, and D.~Huang.
\newblock {CurriculumNet: Weakly Supervised Learning from Large-Scale Web
  Images}.
\newblock In {\em European Conference on Computer Vision ({ECCV})}, 2018.

\bibitem{He2016}
K.~He, X.~Zhang, S.~Ren, and J.~Sun.
\newblock Deep residual learning for image recognition.
\newblock In {\em {IEEE} Conference on Computer Vision and Pattern Recognition
  (CVPR)}, 2016.

\bibitem{Ioffe2015}
S.~Ioffe and C.~Szegedy.
\newblock {Batch Normalization: Accelerating Deep Network Training by Reducing
  Internal Covariate Shift}.
\newblock In {\em International Conference on Machine Learning (ICML)}, 2015.

\bibitem{Jensen2007}
F.~V. Jensen and T.~D. Nielsen.
\newblock {\em Bayesian Networks and Decision Graphs}.
\newblock 2007.

\bibitem{Jiang2015}
L.~Jiang, D.~Meng, Q.~Zhao, S.~Shan, and A.~G. Hauptmann.
\newblock {Self-Paced Curriculum Learning.}
\newblock In {\em Conference on Artificial Intelligence (AAAI)}, 2015.

\bibitem{Jiang2018}
L.~Jiang, Z.~Zhou, T.~Leung, L.-J. Li, and L.~Fei-Fei.
\newblock {{M}entor{N}et: Learning Data-Driven Curriculum for Very Deep Neural
  Networks on Corrupted Labels}.
\newblock In {\em International Conference on Machine Learning (ICML)}, 2018.

\bibitem{Joulin2016}
A.~Joulin, L.~van~der Maaten, A.~Jabri, and N.~Vasilache.
\newblock Learning visual features from large weakly supervised data.
\newblock In {\em European Conference on Computer Vision ({ECCV})}, 2016.

\bibitem{Kapoor2012}
A.~Kapoor, R.~Viswanathan, and P.~Jain.
\newblock Multilabel classification using bayesian compressed sensing.
\newblock In {\em Advances in Neural Information Processing Systems (NIPS)},
  2012.

\bibitem{Kendall2017}
A.~Kendall and Y.~Gal.
\newblock What uncertainties do we need in bayesian deep learning for computer
  vision?
\newblock In {\em Advances in Neural Information Processing Systems (NIPS)},
  2017.

\bibitem{Kingma2014}
D.~P. Kingma and M.~Welling.
\newblock {Auto-Encoding Variational Bayes}.
\newblock In {\em International Conference on Learning Representations (ICLR)},
  2014.

\bibitem{Kornblith2018}
S.~Kornblith, J.~Shlens, and Q.~V. Le.
\newblock {Do Better ImageNet Models Transfer Better?}
\newblock 2018.

\bibitem{Kumar2010}
M.~P. Kumar, B.~Packer, and D.~Koller.
\newblock Self-paced learning for latent variable models.
\newblock In {\em Advances in Neural Information Processing Systems (NIPS)},
  2010.

\bibitem{Kuznetsova2018}
A.~Kuznetsova, H.~Rom, N.~Alldrin, J.~Uijlings, I.~Krasin, J.~Pont-Tuset,
  S.~Kamali, S.~Popov, M.~Malloci, T.~Duerig, and V.~Ferrari.
\newblock {The Open Images Dataset V4: Unified image classification, object
  detection, and visual relationship detection at scale}.
\newblock 2018.

\bibitem{Li2007}
L.~J. Li, G.~Wang, and L.~Fei-Fei.
\newblock {OPTIMOL: automatic Online Picture collecTion via Incremental MOdel
  Learning}.
\newblock In {\em {IEEE} Conference on Computer Vision and Pattern Recognition
  (CVPR)}, 2007.

\bibitem{Li2017c}
W.~Li, L.~Wang, W.~Li, E.~Agustsson, J.~Berent, A.~Gupta, R.~Sukthankar, and
  L.~Van~Gool.
\newblock {WebVision Challenge: Visual Learning and Understanding With Web
  Data}.
\newblock In {\em arXiv 1705.05640}, 2017.

\bibitem{Li2017b}
Y.~Li, Y.~Song, and J.~Luo.
\newblock {Improving Pairwise Ranking for Multi-label Image Classification}.
\newblock In {\em {IEEE} Conference on Computer Vision and Pattern Recognition
  (CVPR)}, 2017.

\bibitem{Lin2014}
T.-Y. Lin, M.~Maire, S.~Belongie, L.~Bourdev, R.~Girshick, J.~Hays, P.~Perona,
  D.~Ramanan, C.~L. Zitnick, and P.~Dollár.
\newblock {Microsoft COCO: Common Objects in Context}.
\newblock 2014.

\bibitem{Liu2018}
C.~Liu, B.~Zoph, J.~Shlens, W.~Hua, L.-J. Li, L.~Fei-Fei, A.~Yuille, J.~Huang,
  and K.~Murphy.
\newblock {Progressive Neural Architecture Search}.
\newblock In {\em European Conference on Computer Vision ({ECCV})}, 2018.

\bibitem{Mahajan2018}
D.~Mahajan, R.~Girshick, V.~Ramanathan, K.~He, M.~Paluri, Y.~Li, A.~Bharambe,
  and L.~van~der Maaten.
\newblock {Exploring the Limits of Weakly Supervised Pretraining}.
\newblock In {\em European Conference on Computer Vision ({ECCV})}, 2018.

\bibitem{Chen2013b}
{Minmin Chen and Alice Zheng and Kilian Weinberger}.
\newblock Fast image tagging.
\newblock In {\em International Conference on Machine Learning (ICML)}, 2013.

\bibitem{Mitchell2015}
T.~Mitchell, W.~Cohen, E.~Hruschka, P.~Talukdar, J.~Betteridge, A.~Carlson,
  B.~Dalvi, M.~Gardner, B.~Kisiel, J.~Krishnamurthy, N.~Lao, K.~Mazaitis,
  T.~Mohamed, N.~Nakashole, E.~Platanios, A.~Ritter, M.~Samadi, B.~Settles,
  R.~Wang, D.~Wijaya, A.~Gupta, X.~Chen, A.~Saparov, M.~Greaves, and
  J.~Welling.
\newblock {Never-Ending Learning}.
\newblock In {\em Conference on Artificial Intelligence (AAAI)}, 2015.

\bibitem{Oquab2014}
M.~Oquab, L.~Bottou, I.~Laptev, and J.~Sivic.
\newblock {Learning and Transferring Mid-Level Image Representations using
  Convolutional Neural Networks}.
\newblock In {\em {IEEE} Conference on Computer Vision and Pattern Recognition
  (CVPR)}, 2014.

\bibitem{Oquab2015}
M.~Oquab, L.~Bottou, I.~Laptev, and J.~Sivic.
\newblock {Is Object Localization for Free? - Weakly-Supervised Learning With
  Convolutional Neural Networks}.
\newblock In {\em {IEEE} Conference on Computer Vision and Pattern Recognition
  (CVPR)}, 2015.

\bibitem{Paszke2017}
A.~Paszke, S.~Gross, S.~Chintala, G.~Chanan, E.~Yang, Z.~DeVito, Z.~Lin,
  A.~Desmaison, L.~Antiga, and A.~Lerer.
\newblock Automatic differentiation in pytorch.
\newblock In {\em Advances in Neural Information Processing Systems (NIPS)},
  2017.

\bibitem{Pham2018}
H.~Pham, M.~Y. Guan, B.~Zoph, Q.~V. Le, and J.~Dean.
\newblock {Faster Discovery of Neural Architectures by Searching for Paths in a
  Large Model}.
\newblock In {\em International Conference on Learning Representations (ICLR)},
  2018.

\bibitem{Qi2017}
X.~Qi, R.~Liao, J.~Jia, S.~Fidler, and R.~Urtasun.
\newblock {3D Graph Neural Networks for RGBD Semantic Segmentation}.
\newblock In {\em {IEEE} International Conference on Computer Vision (ICCV)},
  2017.

\bibitem{Rijsbergen1979}
C.~J.~V. Rijsbergen.
\newblock {\em Information Retrieval}.
\newblock 1979.

\bibitem{Russakovsky2015}
O.~Russakovsky, J.~Deng, H.~Su, J.~Krause, S.~Satheesh, S.~Ma, Z.~Huang,
  A.~Karpathy, A.~Khosla, M.~Bernstein, A.~C. Berg, and L.~Fei-Fei.
\newblock {ImageNet} large scale visual recognition challenge.
\newblock {\em International Journal of Computer Vision (IJCV)}, 2015.

\bibitem{Scarselli2009}
F.~Scarselli, M.~Gori, A.~C. Tsoi, M.~Hagenbuchner, and G.~Monfardini.
\newblock {The graph neural network model}.
\newblock {\em IEEE Transactions on Neural Networks}, 2009.

\bibitem{Stock2018}
P.~Stock and M.~Cisse.
\newblock {ConvNets and ImageNet Beyond Accuracy: Understanding Mistakes and
  Uncovering Biases}.
\newblock In {\em European Conference on Computer Vision ({ECCV})}, 2018.

\bibitem{Sun2016}
C.~Sun, M.~Paluri, R.~Collobert, R.~Nevatia, and L.~Bourdev.
\newblock {ProNet: Learning to Propose Object-Specific Boxes for Cascaded
  Neural Networks}.
\newblock In {\em {IEEE} Conference on Computer Vision and Pattern Recognition
  (CVPR)}, 2016.

\bibitem{Sun2017}
C.~Sun, A.~Shrivastava, S.~Singh, and A.~Gupta.
\newblock {Revisiting Unreasonable Effectiveness of Data in Deep Learning Era}.
\newblock In {\em {IEEE} International Conference on Computer Vision (ICCV)},
  2017.

\bibitem{Sun2010}
Y.-Y. Sun, Y.~Zhang, and Z.-H. Zhou.
\newblock {Multi-label Learning with Weak Label}.
\newblock In {\em Conference on Artificial Intelligence (AAAI)}, 2010.

\bibitem{Szegedy2017}
C.~Szegedy, S.~Ioffe, V.~Vanhoucke, and A.~Alemi.
\newblock Inception-v4, inception-resnet and the impact of residual connections
  on learning.
\newblock In {\em Conference on Artificial Intelligence (AAAI)}, 2017.

\bibitem{Tang2009}
L.~Tang, S.~Rajan, and V.~K. Narayanan.
\newblock {Large scale multi-label classification via metalabeler}.
\newblock In {\em WWW}, 2009.

\bibitem{Tsoumakas2007}
G.~Tsoumakas and I.~Katakis.
\newblock Multi-label classification: An overview.
\newblock {\em International Journal of Data Warehousing and Mining (IJDWM)},
  2007.

\bibitem{Vahdat2017}
A.~Vahdat.
\newblock Toward robustness against label noise in training deep discriminative
  neural networks.
\newblock In {\em Advances in Neural Information Processing Systems (NIPS)},
  2017.

\bibitem{Vasisht2014}
D.~Vasisht, A.~Damianou, M.~Varma, and A.~Kapoor.
\newblock {Active Learning for Sparse Bayesian Multilabel Classification}.
\newblock In {\em ACM SIGKDD International Conference on Knowledge Discovery
  and Data Mining}, 2014.

\bibitem{Wang2014}
Q.~Wang, B.~Shen, S.~Wang, L.~Li, and L.~Si.
\newblock {Binary Codes Embedding for Fast Image Tagging with Incomplete
  Labels}.
\newblock In {\em European Conference on Computer Vision ({ECCV})}, 2014.

\bibitem{Wu2015}
B.~Wu, S.~Lyu, and B.~Ghanem.
\newblock {ML-MG: Multi-Label Learning With Missing Labels Using a Mixed
  Graph}.
\newblock In {\em {IEEE} International Conference on Computer Vision (ICCV)},
  2015.

\bibitem{Xie2017}
S.~Xie, R.~Girshick, P.~Dollár, Z.~Tu, and K.~He.
\newblock {Aggregated Residual Transformations for Deep Neural Networks}.
\newblock In {\em {IEEE} Conference on Computer Vision and Pattern Recognition
  (CVPR)}, 2017.

\bibitem{Xu2013}
M.~Xu, R.~Jin, and Z.-H. Zhou.
\newblock {Speedup Matrix Completion with Side Information: Application to
  Multi-Label Learning}.
\newblock In {\em Advances in Neural Information Processing Systems (NIPS)},
  2013.

\bibitem{Yang2016}
H.~Yang, J.~T. Zhou, and J.~Cai.
\newblock {Improving Multi-label Learning with Missing Labels by Structured
  Semantic Correlations}.
\newblock In {\em European Conference on Computer Vision ({ECCV})}, 2016.

\bibitem{Yang1999}
Y.~Yang.
\newblock An evaluation of statistical approaches to text categorization.
\newblock 1999.

\bibitem{Yu2014}
H.-F. Yu, P.~Jain, P.~Kar, and I.~S. Dhillon.
\newblock {Large-scale Multi-label Learning with Missing Labels}.
\newblock In {\em International Conference on Machine Learning (ICML)}, 2014.

\bibitem{Zhang2017}
C.~Zhang, S.~Bengio, M.~Hardt, B.~Recht, and O.~Vinyals.
\newblock {Understanding deep learning requires rethinking generalization}.
\newblock In {\em International Conference on Learning Representations (ICLR)},
  2017.

\bibitem{Zhou2016}
B.~Zhou, A.~Khosla, A.~Lapedriza, A.~Oliva, and A.~Torralba.
\newblock {Learning Deep Features for Discriminative Localization}.
\newblock In {\em {IEEE} Conference on Computer Vision and Pattern Recognition
  (CVPR)}, 2016.

\bibitem{Zoph2018}
B.~Zoph, V.~Vasudevan, J.~Shlens, and Q.~V. Le.
\newblock Learning transferable architectures for scalable image recognition.
\newblock In {\em {IEEE} Conference on Computer Vision and Pattern Recognition
  (CVPR)}, 2018.

\end{thebibliography}
